\documentclass[journal]{IEEEtran}

\usepackage{amsmath}
\usepackage{amssymb}
\usepackage{amsfonts}
\usepackage{pifont}
\usepackage{amsthm}
\usepackage{bbm}
\usepackage{enumitem}
\usepackage{subfiles}
\usepackage{xr}
\usepackage{algorithmic}
\usepackage{graphicx}
\usepackage{subfig}
\usepackage{cite}
\usepackage{hyperref}
\usepackage{xcolor}

\newtheorem{theorem}{Theorem}[section]

\newtheorem{lemma}[theorem]{Lemma}

\DeclareMathOperator*{\argmin}{arg\,min}
\DeclareMathOperator{\spn}{span}

\newcommand{\cmark}{\text{\ding{51}}}
\newcommand{\xmark}{\text{\ding{55}}}

\begin{document}
\title{Modularizing Deep Learning via Pairwise Learning With Kernels}

\author{Shiyu~Duan, Shujian~Yu,~\IEEEmembership{Member,~IEEE}, and Jos\'{e}~C.~Pr\'{i}ncipe,~\IEEEmembership{Life~Fellow,~IEEE}
\thanks{SD (\href{mailto:michaelshiyu3@gmail.com}{michaelshiyu3@gmail.com}) and JCP (\href{mailto:principe@cnel.ufl.edu}{principe@cnel.ufl.edu}) are with the Department of Electrical and Computer Engineering, University of Florida, Gainesville, FL, 32611. SY (\href{mailto:yusj9011@gmail.com}{yusj9011@gmail.com}) is with NEC Labs Europe, 69115 Heidelberg, Germany.}
\thanks{Code available at: \url{https://github.com/michaelshiyu/kerNET.git}.}}

\markboth{Preprint. Under review.}%
{Duan, Yu, and Pr\'{i}ncipe: Modularizing Deep Learning via Pairwise Learning With Kernels}

\maketitle

\begin{abstract}
    By redefining the conventional notions of layers, we present an alternative view on finitely wide, fully trainable deep neural networks as stacked linear models in feature spaces, leading to a kernel machine interpretation.
    Based on this construction, we then propose a provably optimal modular learning framework for classification that does not require between-module backpropagation.
    This modular approach brings new insights into the label requirement of deep learning:
    It leverages only implicit pairwise labels (weak supervision) when learning the hidden modules.
    When training the output module, on the other hand, it requires full supervision but achieves high label efficiency, needing as few as 10 randomly selected labeled examples (one from each class) to achieve 94.88\% accuracy on CIFAR-10 using a ResNet-18 backbone.
    Moreover, modular training enables fully modularized deep learning workflows, which then simplify the design and implementation of pipelines and improve the maintainability and reusability of models.
    To showcase the advantages of such a modularized workflow, we describe a simple yet reliable method for estimating reusability of pre-trained modules as well as task transferability in a transfer learning setting.
    At practically no computation overhead, it precisely described the task space structure of 15 binary classification tasks from CIFAR-10.
\end{abstract}



\section{Introduction}
\IEEEPARstart{U}{nderstanding} the connections between neural networks (NNs) and kernel methods has been a long-standing goal in machine learning research~\cite{neal1995bayesian}.
Recently, there has been a resurgence of interest in this direction, leading to important insights and powerful algorithms~\cite{lee2017deep,jacot2018neural,shankar2020neural,cho2009kernel,arora2019harnessing,arora2019exact,li2019enhanced}.
The established connections require highly nontrivial assumptions: The equivalence between a particular kernel method and a family of NNs only exists in infinitely wide NNs and/or in expectation of random NNs (only weakly-trained~\cite{arora2019exact}).
Moreover, existing algorithms using kernels inspired by NNs typically have prohibitively high computational complexity (super-quadratic in sample size, to be exact~\cite{arora2019harnessing}).

\begin{figure}[!t]
    \centering
    \includegraphics[width=\columnwidth]{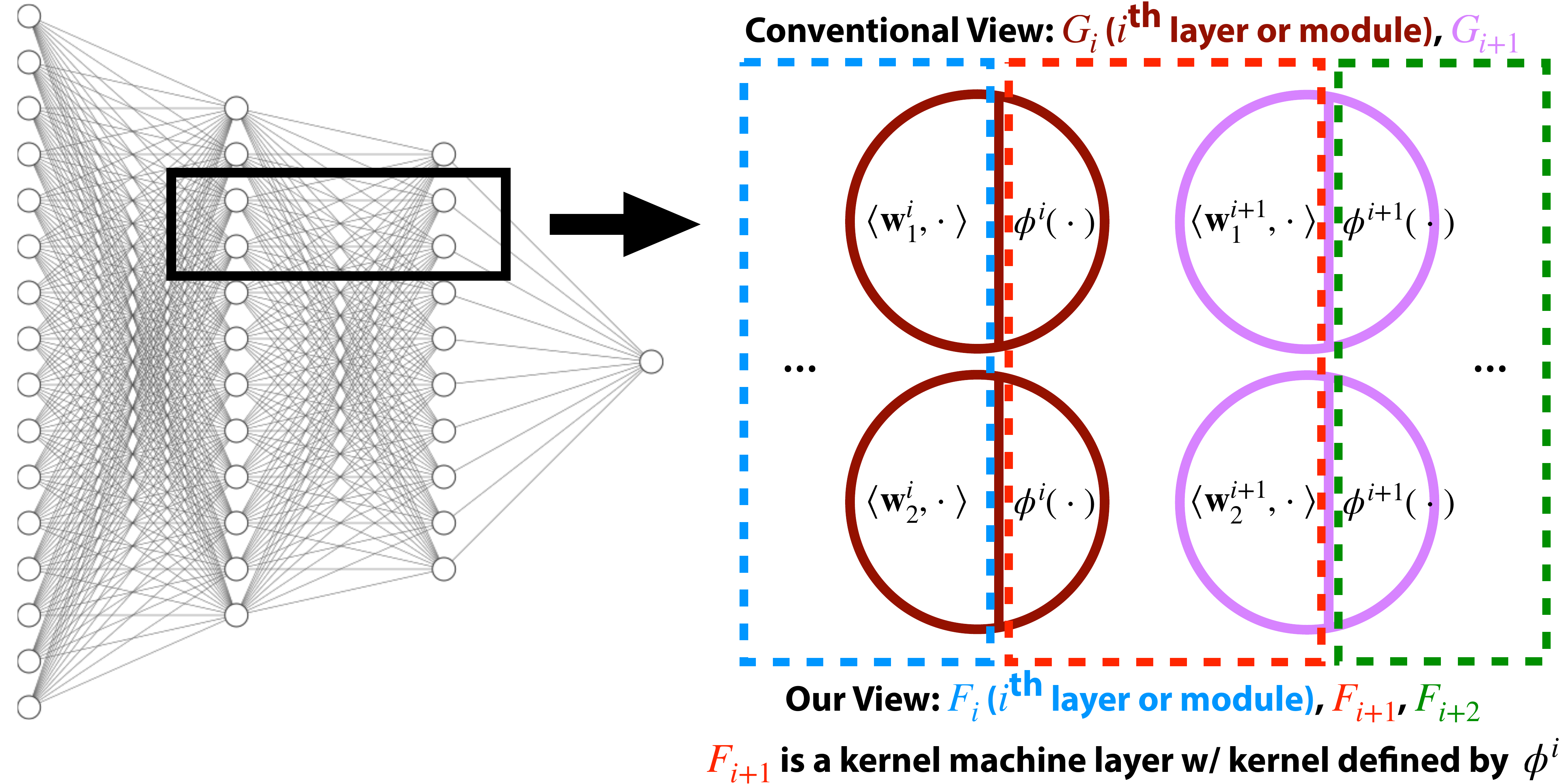}
    \caption{
        Viewing the models from a new perspective, we identify the kernel machines ``hidden'' in neural networks.
        Specifically, by absorbing the trailing nonlinearity of a layer or a module (composition of layers) into the next layer, layers become linear models in feature spaces.
        These linear models can be shown to be kernel machines.
        Best viewed in color.
    }
    \label{fig1}
\end{figure}

In this paper, we propose a simple method to link fully trainable, finitely wide NNs to kernel machines (KMs), i.e., linear models in feature spaces (reproducing kernel Hilbert spaces (RKHSs)) (Fig.~\ref{fig1}).
Specifically, as opposed to the existing literature, where the nonlinearity is considered the \textit{last} component of an NN layer or module (composition of layers), we consider it to be the \textit{first} component of the next layer.
Then layers in deep NN can be identified as linear models in feature spaces connected by potentially nonlinear feature maps.
These linear models can be proven to be KMs with kernels induced by the connecting feature maps.
The benefits of our construction are three-fold: First, the connections are established for fully trainable and finitely wide NNs, and are exact in the sense that the trainable parameters of the NN layers are exactly those of the corresponding KMs.
Second, our method works with any NN layer, fully-connected or convolutional, with minimal adjustments.
Last but not least, these NN-equivalent KMs run in linear time, contrasting the super-quadratic runtime of existing algorithms.


We then turn to another important yet understudied problem: How can we modularize deep learning (DL)?
In engineering and software engineering in particular, modularization is the core of any scalable workflow.
Dividing the design into modules, optimizing them individually, and then wiring them together simplifies implementation via enabling unit tests, and enhances maintainability and reusability.
There is currently no reliable way to completely modularize a deep NN, however, since there is no modular learning approach that provably matches the performance of end-to-end training.
As a result, one is forced to design and optimize the entire model as a whole, configuring hundreds of hyperparameters simultaneously.
When performance is unsatisfying, it is practically impossible to trace the source of the problem to a particular design choice.
Even if one succeeded in training a good model, reusing that model for a new task is highly nontrivial: Which part of the model is more reusable for what task?

To enable fully modularized DL, we develop a provably optimal modular training framework for deep NNs in classification that trains each module separately, yet still finds the overall loss function minimizer as an end-to-end approach would.
Focusing on the two-module case, we prove that the training of input and output modules can be decoupled by leveraging pairwise kernel evaluations on training examples from distinct classes.
This kernel is defined by the output module's nonlinearity.
It suffices to have the input module optimize a proxy objective function that does not involve the trainable parameters of the output one.
This removes the need for error backpropagation between modules.
On MNIST and CIFAR-10, our modular approach compares favorably against end-to-end training in terms of accuracy.

Our modular learning utilizes labels more efficiently than the existing end-to-end training paradigm.
Specifically, training of the input module involves only pairs of examples from distinct classes with no need for knowing the actual classes.
This is a weaker form of supervision than knowing exactly which class each example belongs to as required by backpropagation. 
And we empirically show that the output module, which requires full supervision in training, is highly label-efficient, achieving \(94.88\%\) accuracy on CIFAR-10 with \(10\) randomly selected labeled examples (one from each class) using a ResNet-18~\cite{he2016deep} backbone (\(94.93\%\) when using all \(50000\) labels).
Overall, our modular training can leverage a weaker form of supervision yet still produce models as performant as those obtained from end-to-end backpropagation, indicating that the existing form of supervision and backpropagation is not efficient enough.
This can potentially enable less costly procedures for acquiring labeled data and more powerful un/semi-supervised learning algorithms.

To showcase one of the main benefits of modularization --- module reuse with confidence --- we demonstrate that one can easily and reliably quantify the reusability of a pre-trained module on a new target task with our proxy objective function.
This provides a fast yet effective solution to an important practical issue in transfer learning.
Moreover, this method can be extended to measure task transferability, a central problem in transfer learning, continual/lifelong learning, and multi-task learning~\cite{tran2019transferability}.
Unlike many existing methods, our approach requires no training, is task agnostic, flexible, and completely data-driven.
Nevertheless, it accurately described the task space structure on \(15\) binary classification tasks derived from CIFAR-10 using only a small amount of labeled data.


To summarize, our main contributions are as follows.
\begin{itemize}
    \item We present a simple method that establishes connections between fully trainable, finitely wide NNs and KMs, contrasting earlier efforts that always assume random networks or require infinite widths. (Sec.~\ref{sec1})
    \item We present a modular training framework, opening up new possibilities for modularized DL.
    Focusing on the two-module case, we provide an optimality proof for our learning approach, guaranteeing that it finds the loss function minimizer without between-module backpropagation.
    As empirical validation, we demonstrate its strong performance on MNIST and CIFAR-10. (Sec.~\ref{sec2})
    \item Our modular learning approach sheds new light on the label requirement of DL: It relies almost only on implicitly-labeled example pairs, achieving state-of-the-art accuracy on CIFAR-10 with a single randomly chosen fully-labeled example per class. (Sec.~\ref{sec4})
    \item We propose a simple but effective method to quantify module reusability and task transferability using components from our modular training framework, demonstrating that modularized DL enables solutions to important issues in transfer learning, lifelong learning, etc. (Sec.~\ref{sec3})
\end{itemize}

\section{Notations and Background}

\subsection{Notations}
Throughout, we use bold capital letters for matrices and tensors, bold lower-case letters for vectors, and unbold letters for scalars.
\((\mathbf{v})_i\) denotes the \(i^\text{th}\) component of vector \(\mathbf{v}\).
And \(\mathbf{W}^{(j)}\) denotes the \(j^\text{th}\) column of matrix \(\mathbf{W}\).
For a 3D tensor \(\mathbf{X}\), \(\mathbf{X}[:,:,c]\) denotes the \(c^\text{th}\) matrix indexing along the third dimension from the left (or the \(c^\text{th}\) channel).
We use \(\langle \cdot, \cdot \rangle_H\) to denote the inner product in an inner product space \(H\).
And the subscript shall be omitted if doing so causes no confusion.
For functions, we use capital letters to denote vector/matrix/tensor-valued functions, and lower-case letters are reserved specifically for scalar-valued ones.
In a network, we call a composition of an arbitrary number of layers as a module for convenience.

Since we shall propose an alternative view on NNs that redefine the network layers and also modules, it is helpful to introduce notations to distinguish our view and the conventional one.
We use letter \(G_i\) (or \(g_i\), depending on if this layer or module is a scalar-valued function) with a numeric subscript \(i\in\mathbb{N}\setminus\{0\}\) to refer to the \(i^\text{th}\) network layer or module under the conventional view and letter \(F_i\) (or \(f_i\)) the \(i^\text{th}\) layer or module (of the same model) under our view.
Example usages can be found in Fig.~\ref{fig1} and Sec.~\ref{sec1}.


\subsection{Kernels and Kernel Machines}
Kernel machines can be considered as linear models in feature spaces, the mappings into which are potentially nonlinear~\cite{shalev2014understanding}.
Consider a feature map \(\Phi: \mathbb{R}^d \to H\), where \(H\) is a (real) Hilbert space, i.e., an inner product space over the real numbers that is complete in the metric induced by the inner product, a kernel machine is a linear model in this feature space \(H\):
\begin{equation}
    f(\mathbf{x}) = \langle \mathbf{w}, \Phi(\mathbf{x}) \rangle_H + b, \mathbf{w}\in H, b\in\mathbb{R},
    \label{eq1}
\end{equation}
where \(\mathbf{w}, b\) are its trainable weights and bias, respectively.
For certain feature spaces \(H\) (called RKHSs), one can define a bivariate ``kernel'' function \(k\)  that is symmetric and positive semidefinite via
\begin{equation}
    k(\mathbf{u}, \mathbf{v}) := \langle \Phi(\mathbf{u}), \Phi(\mathbf{v})\rangle_H.
    \label{eq2}
\end{equation}
This definition is often used as an identity and is referred to as the ``kernel trick'' in some more modern texts~\cite{shalev2014understanding}.\footnote{The correspondence between \(k\) and \(\Phi\) holds true in both directions and is sometimes stated the other way around. Namely, per Moore-Aronszajn Theorem, for every symmetric, continuous, positive semidefinite bivariate function \(k\) that maps into \(\mathbb{R}\), one can find an RKHS \(H\) such that \(k(\mathbf{u}, \mathbf{v}) = \langle \Phi(\mathbf{u}), \Phi(\mathbf{v})\rangle_H\), where the feature map \(\Phi\) is defined through \(k\) as \(\Phi(\mathbf{u}) := k(\mathbf{u}, \cdot)\)~\cite{aronszajn1950theory}.}

An RKHS, different from a general Hilbert space, possesses a ``canonical coordinate system'' induced by the kernel~\cite{manton2015primer} (in addition to the regular coordinate system induced by its basis) that enables one to concisely express distance between feature vectors using kernel values, thanks to the reproducing property~\cite{scholkopf2001learning}.
Concretely, we have
\begin{equation}
    \|\Phi(\mathbf{u}) - \Phi(\mathbf{v})\|_H^2 = k(\mathbf{u}, \mathbf{u}) + k(\mathbf{v}, \mathbf{v}) - k(\mathbf{u}, \mathbf{v}).
\end{equation}

One can use the ``kernel trick'' to implement highly nontrivial feature maps (thus highly complicated functions in the input space).
For feature maps that are not implementable, e.g., when \(\Phi(\mathbf{x})\) is an infinite series, one can approximate the KM with a set of ``centers'' \(\mathbf{x}_1, \ldots, \mathbf{x}_n\) as follows:
\begin{equation}
    f(\mathbf{x}) \approx \langle \mathbf{w}, \Phi(\mathbf{x}) \rangle_H + b, \mathbf{w}\in \spn\{\Phi(\mathbf{x}_1), \ldots, \Phi(\mathbf{x}_n)\}, b\in\mathbb{R},
\end{equation}
Assuming a \(k\) satisfying Eq.~\ref{eq2} can be found and using the ``kernel trick'', the right hand side is equal to
\begin{equation}
    \sum_i^n \alpha_i k(\mathbf{x}_i, \mathbf{x}) + b, \alpha_i, b\in\mathbb{R}.
\end{equation}
Now, the learnable parameters become the \(\alpha_i\)'s anb \(b\).


This approximation changes the computational complexity of evaluating the kernel machine on a single example from \(\mathcal{O}(h)\), where \(h\) is the dimension of the feature space \(H\) and can be infinite for certain kernels, to \(\mathcal{O}(nd)\), where \(d\) is the dimension of the input space.
In practice, the runtime for a sample is quadratic in sample size since \(n\) is typically on the same order as the sample size.
There exists acceleration methods that reduce the complexity via further approximation in this case (e.g.,~\cite{rahimi2008random}), yet the compromise in performance can be nonnegligible in practice especially when the input space dimension is large.


\section{Revealing the Hidden Kernel Machines in Neural Networks}
\label{sec1}
In this section, we present a method revealing the hidden KMs in NNs.
The idea is simple: Instead of considering the nonlinearity to be the \textit{last} component of a layer or a module (call it the \(i^\text{th}\) layer or module here for convenience), we consider it to be the \textit{first} component of the next layer, i.e., layer \(i + 1\).
After potentially repeating this process for layer \(i + 1\) to get rid of its trailing nonlinearity, we turned layer \(i + 1\) into a layer of linear models in feature spaces.
These linear models can be shown to be KMs (Fig.~\ref{fig1}).
We first present the method for fully-connected networks and then extend to CNNs~\cite{lecun1998gradient}.
Contrasting existing works (e.g., \cite{arora2019exact}), the extension requires minimal adjustments only.

\subsection{The Methodology}
\label{static}
\subsubsection{Fully-Connected Neural Networks}
We use a one-hidden-layer NN as an illustrative example.
The idea scales easily to deeper models.
Note that we assume the output layer has a single neuron.
When the output layer has multiple neurons, one can apply the same analysis to each of these output neurons.
Finally, we assume the output layer to be linear without loss of generality since if there is a trailing nonlinearity at the output, it can be viewed as a part of the loss function instead of a part of the network.

Considering an input vector \(\mathbf{x} \in \mathbb{R}^{d_0}\), the model \(f = g_2\circ G_1\) is given as
\begin{align}
    &G_1(\mathbf{x}) = \Phi(\mathbf{W}_1^\top \mathbf{x}) \in \mathbb{R}^{d_1}; \\
    &f(\mathbf{x}) = g_2(G_1(\mathbf{x})) = \mathbf{w}_2^\top G_1(\mathbf{x}),
\end{align}
where \(\Phi\) is an elementwise nonlinearity such as ReLU~\cite{nair2010rectified} with \(\Phi(\mathbf{v}) := \left( \phi\left( (\mathbf{v})_1 \right), \ldots,\phi\left( (\mathbf{v})_{d_1} \right) \right)^\top\) for any \(\mathbf{v}\in\mathbb{R}^{d_1}\) and some \(\phi:\mathbb{R}\to\mathbb{R}\), \(G_1\) the input layer, \(g_2\) the output layer, \(f(\mathbf{x})\) the final model output, and \(\mathbf{W}_1\) and \(\mathbf{w}_2\) the learnable weights of the NN model.

There are different ways of dividing this particular model into ``layers''.
Indeed, without changing the input-output mapping, we can redefine the layers as follows.
\begin{align}
    f(\mathbf{x}) = \langle \mathbf{w}_2, \Phi(\mathbf{W}_1^\top\mathbf{x})\rangle_{\mathbb{R}^{d_1}}
    = \langle \mathbf{w}_2, \Phi(F_1(\mathbf{x}))\rangle_{\mathbb{R}^{d_1}} = f_2(F_1(\mathbf{x})),
\end{align}
where \(F_1(\mathbf{x}) = \left(\langle \mathbf{W}_1^{(1)}, \mathbf{x}\rangle_{\mathbb{R}^{d_0}}, \ldots, \langle \mathbf{W}_1^{(d_1)}, \mathbf{x}\rangle_{\mathbb{R}^{d_0}}\right)^\top\) and \(\langle\cdot, \cdot\rangle_{\mathbb{R}^k}\) is the canonical inner product of \(\mathbb{R}^k\) for \(k = d_0, d_1\), i.e., the dot product.
In other words, we treat \(F_1\) as the new input layer and absorb the nonlinearity \(\Phi\) into the old output layer, forming the new output layer \(f_2\) such that each layer is now one or multiple parallel linear models (models linear in their weights, not necessarily in their inputs) (Fig.~\ref{fig1}).

Then one-hidden layer NN can be interpreted as a composition of KMs with finite-dimensional RKHSs.
Using Eq.~\ref{eq2}, the linear models on the input layer are simply KMs with feature maps being the identity map on \(\mathbb{R}^{d_0}\), inducing the identity kernel, i.e., \(k_I(\mathbf{u}, \mathbf{v}) := \langle \mathbf{u}, \mathbf{v}\rangle_{\mathbb{R}^{d_0}}\).
On the other hand, the linear model on the output layer, i.e., \(f_2\), can be easily identified as a KM with feature map being \(\Phi\), (and together with the input layer) inducing kernel \(k\left(F_1(\mathbf{u}), F_1(\mathbf{v})\right) := \left\langle\Phi\left(F_1(\mathbf{u})\right), \Phi\left(F_1(\mathbf{v})\right)\right\rangle_{\mathbb{R}^{d_1}}\).

\subsubsection{Convolutional Neural Networks}
\begin{figure}[!t]
    \centering
    \includegraphics[width=\columnwidth]{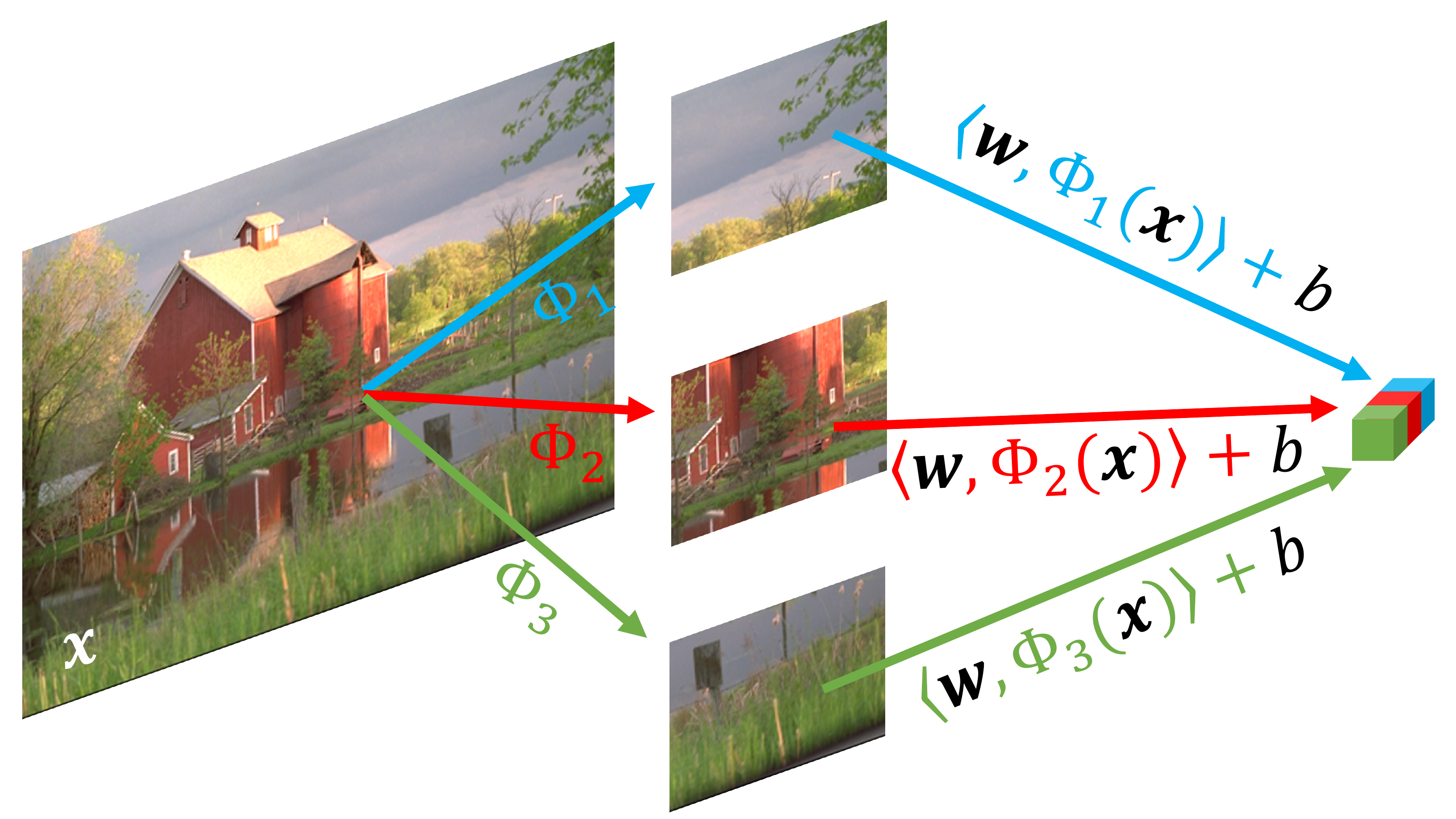}
    \caption{
        A convolutional layer (illustrated is a single-filter instantiation) is multiple kernel learning: It can be equated to concatenated KMs using distinct kernels but sharing weights.
        Each color corresponds to a KM.
        Elements in black are shared across KMs.
        Best viewed in color.
    }
    \label{fig2}
\end{figure}
The above method for fully-connected networks can be extended to CNNs as follows.
We discuss here simple 2D convolutional networks and the same idea generalizes to more complicated models easily.

Similar to the fully-connected case, we first absorb the trailing elementwise nonlinearity \(\Phi\) of each layer or module of interest into the next layer.
Suppose the activation tensor of a given layer or module is \(\mathbf{X}\in\mathbb{R}^{H}\times \mathbb{R}^W\times \mathbb{R}^C\), then each channel of the next layer is a matrix of KMs sharing the same weights with the \(ij^\text{th}\) KM having feature map: \(\Phi_{ij}(\mathbf{X}) := r_{ij}\circ\Phi(\mathbf{X})\), where \(r_{ij}:\mathbb{R}^{H}\times \mathbb{R}^W\times \mathbb{R}^C\to\mathbb{R}^{h\times w\times C}:\mathbf{Z}\mapsto [s_{ij}(\mathbf{Z}[:,:,1]), ..., s_{ij}(\mathbf{Z}[:,:,C])]\), \(s_{ij}\) denotes the operator that returns a \(h\times w\) vectorized receptive field centered at a specific location (depending on \(ij\)) upon receiving a matrix, and \([\cdot, ..., \cdot]\) denotes vector concatenation.
This is illustrated in Fig.~\ref{fig2}.

There are two main differences compared to the fully-connected case.
First, in the convolutional case, the KMs on each channel in a given layer share the same weights whereas in the fully-connected case, there is no such restriction.
Second, these KMs have \textit{distinct} kernel functions, unlike in the fully-connected case where they share kernels.
These observations allow for a new perspective in interpreting CNNs.
Indeed, the fact that each convolutional layer is essentially KMs using distinct kernels but sharing weights suggests that it can be viewed as an instantiation of the multiple kernel learning framework that has been studied in the kernel method literature for decades~\cite{gonen2011multiple}.
However here the composition is different because it is an embedding of functions. 


%

\subsection{Implementable Feature Maps and Linear Runtime}
Note that, for common NNs, the kernels involved in the above methods have implementable feature maps with feature space dimensions being equal to the widths of the corresponding NN layers (so they are always finite).
Therefore, one no longer needs to rely on the kernel trick and approximate the kernel machines through pairwise evaluations on a set of centers --- one can simply use the explicit linear model representation \(f(\mathbf{x}) = \langle \mathbf{w}, \Phi(\mathbf{x})\rangle + b\)!
The major advantage is that the computational complexity of running the model on a single data example becomes \(\mathcal{O}(p)\), where \(p\) is the size of the corresponding NN layer (or equivalently, the size of \(\Phi(\mathbf{x})\)).
This reduces the runtime over a set of data from the usual super-quadratic to linear in sample size, making the model much more practical.

\section{Provably Optimal Modular Learning}
\label{sec2}
Constructing a unified model class that encompasses both NNs and KMs yields immediately practical implications.
Namely, existing algorithms and results for KMs can now be directly applied to NNs and vice versa.
Specifically, one may expect the theory behind KMs (see, e.g.,~\cite{vapnik2000,scholkopf2001learning,shawe2004kernel}) to significantly enrich our understanding of NNs and enable novel algorithms.

In this section, we propose a provably optimal modular training framework for NNs in classification, which enables fully modularized DL workflows.
We focus on the two-module case.
The idea can be easily generalized to enable modular training with more than two modules by analyzing one pair of modules at a time.
We leave the corresponding optimality proofs as a future work.
Applying this framework to NNs essentially relies on viewing the output module as a KM and utilizing its linearity in an RKHS and properties of the RKHS.
This demonstrates how the interplay of NN and KM theories can produce useful results.

\subsection{Set-Up, Goal, and an Idea}
Suppose we have a deep model consisting of two modules \(F = F_2\circ F_1\) and an objective function \(L(F, S)\), where \(S\) is a training set \(S=\{(\mathbf{x}_i, y_i)\}_{i=1}^n\).
\(F_1, F_2\) can be compositions of arbitrary layers.

A modular learning algorithm trains \(F_1\), freezes it afterwards, then trains \(F_2\).
And the goal is that after wiring together the two trained modules, the overall model minimizes \(L\).

For a given \(S\), define
\begin{equation}
    \label{def}
    \mathcal{F}_1^\star := \{F_1: \exists F_2 \text{ s.t. } F_2\circ F_1\in\argmin_F L(F, S)\}.
\end{equation}
Clearly, to obtain an \(F\) that minimizes \(L\), the goal when training \(F_1\) can be to find an \(F_1^\prime\) that is in \(\mathcal{F}_1^\star\).
Then one can simply train \(F_2\) to minimize \(L(F_2\circ F_1^\prime, S)\) and the resulting minimizer \(F_2^\prime\) will satisfy \(F_2^\prime\circ F_1^\prime\in \argmin_F L(F, S)\).
And if we can characterize \(\mathcal{F}_1^\star\) independently of the trainable parameters of \(F_2\), the training of \(F_1\) can be decoupled from that of \(F_2\).



\subsection{Main Result}

Earlier we have shown how to redefine NN layers such that they become KMs, but we have not yet demonstrated any practical advantage of such a representation.
In this section, we show that under this KM view on the output module, the optimal input module can be described independently of the output module, which, as stated above, is the core idea motivating our modular learning algorithm.
For simplicity, we discuss only binary classification.
The result easily extend to classification with more classes.

Concretely, we present a result stating that if \(F_2\) is a single KM with kernel \(k\), \(\mathcal{F}_1^\star\) can be characterized independently of the trainable parameters of \(F_2\) via pairwise evaluations of the kernel \(k\) on training data from distinct classes.
The proof of the following theorem is given in the Appendix.



    \begin{theorem}
        \label{th1}
        Let \(S = \{(\mathbf{x}_i, y_i)\}_{i=1}^n, \mathbf{x}_i\in\mathbb{R}^{d_0}, y_i\in\{+, -\}, \forall i\), be given and consider \(F_1:\mathbb{R}^{d_0}\to\mathbb{R}^{d_1}, f_2:\mathbb{R}^{d_1}\to\mathbb{R}: \mathbf{z}\mapsto\langle \mathbf{w}, \phi(\mathbf{z}) \rangle + b\), where \(\mathbf{w}, b\) are free parameters and \(\phi\) is a given mapping into a real inner product space with \(\|\phi(\mathbf{u})\| = \alpha\) for all \(\mathbf{u}\in\mathbb{R}^{d_1}\) and some fixed \(\alpha > 0\).\footnote{Throughout, we consider the natural norm induced by the inner product, i.e., \(\|\mathbf{t}\|^2 := \langle \mathbf{t}, \mathbf{t} \rangle, \forall \mathbf{t}\).}
        Let \(I_+\) be the set of \(i\)'s in \(\{1, \ldots, n\}\) such that \(y_i = +\).
        And let \(I_-\) be the set of \(j\)'s in \(\{1, \ldots, n\}\) such that \(y_j = -\).
        Suppose the objective function \(L\) admits the following form:
        \begin{align}
            &L(f_2\circ F_1, S)\nonumber\\
            &\quad= \frac{1}{n}\sum_{i\in I_+}\ell_+\left( f_2\circ F_1(\mathbf{x}_i) \right) + \frac{1}{n}\sum_{j\in I_-}\ell_-\left( f_2\circ F_1(\mathbf{x}_j) \right)\nonumber\\
            &\quad\quad\quad + \lambda g(\|\mathbf{w}\|),
        \end{align}
        where \(\lambda \geq 0\), \(g, \ell_+, \ell_-\) are all real-valued with \(g, \ell_-\)nondecreasing and \(\ell_+\) nonincreasing.

        For an \(F_1^\star\), let \(f_2^\star\) be in \(\argmin_{f_2}L(f_2\circ F_1^\star, S)\).
        If \(\forall i\in I_+, j\in I_-\), \(F_1^\star\) satisfies
        \begin{equation}
            \label{re}
            \|\phi(F_1^\star(\mathbf{x}_i)) - \phi(F_1^\star(\mathbf{x}_j))\| \geq \|\phi(\mathbf{s}) - \phi(\mathbf{t})\|, \forall\mathbf{s}, \mathbf{t}\in\mathbb{R}^{d_1},
        \end{equation}
        then
        \begin{equation}
            f_2^\star\circ F_1^\star\in\argmin_{f_2\circ F_1}L(f_2\circ F_1, S).
        \end{equation}
    \end{theorem}

    \noindent\textbf{Remark:}
    Defining kernel
    \begin{equation}
        k\left(F_1(\mathbf{u}), F_1(\mathbf{v})\right) = \left\langle \phi\left(F_1(\mathbf{u})\right), \phi\left(F_1(\mathbf{v})\right) \right\rangle,
    \end{equation}
    Eq.~\ref{re} is equivalent to
    \begin{equation}
        k\left(F_1^\star(\mathbf{x}_i), F_1^\star(\mathbf{x}_j)\right) \leq k\left( \mathbf{s}, \mathbf{t} \right), \forall\mathbf{s}, \mathbf{t}\in\mathbb{R}^{d_1}, \forall i\in I_+, j\in I_-.
    \end{equation}

    Further, if the infimum of \(k(\mathbf{u}, \mathbf{v})\) is attained in \(\mathbb{R}^{d_1} \times \mathbb{R}^{d_1}\) and equals \(\beta\), then Eq.~\ref{re} is equivalent to
    \begin{equation}
        k\left(F_1^\star(\mathbf{x}_i), F_1^\star(\mathbf{x}_j)\right) =\beta, \forall i\in I_+, j\in I_-.
    \end{equation}

Interpreting a two-module classifier \(F_2\circ F_1\) as \(F_1\) learning a new representation of the given data on which \(F_2\) will carry out the classification, the intuition behind our result can be explained as follows.
Given some data, its ``optimal'' representation for a linear model to classify should be the one where examples from distinct classes are located as distant from each other as possible.
Thus, the optimal \(F_1\) should be the one that produces this optimal representation.
And since our \(F_2\) is a linear model in an RKHS feature space and \textit{because distance in an RKHS can be expressed via evaluations of its reproducing kernel}, the optimal \(F_1\) can then be fully described with only pairwise kernel evaluations over the training data, i.e., \(k\left(F_1(\mathbf{x}_i), F_1(\mathbf{x}_j)\right)\) for \(\mathbf{x}_i, \mathbf{x}_j\) being training examples from different classes.
In other words, an RKHS provides a useful ``canonical coordinate system''~\cite{manton2015primer} that enables a concise definition of the optimal input module for any linear classifier in that RKHS using only kernel evaluations.

\subsection{Applicability of the Main Result}

The earlier Theorem imposes assumptions on the network architecture and the classification objective function.
We now show that these assumptions are satisfied in many standard classification set-ups.

\subsubsection{A Two-Module View on Neural Networks}
\label{two-module view}
Most popular networks, including the ResNet family~\cite{he2016deep}, has the following representation:
\begin{equation}
    F = G_2\circ G_1,
\end{equation}
where \(G_2\) is a output linear layer: \(\left(G_2(\mathbf{x})\right)_i = \langle \mathbf{w}_i, \mathbf{x} \rangle + b_i\) and \(G_1\) is a composition of arbitrary previous layers ending with a nonlinearity \(\Phi\).
When the model also uses a nonlinearity on top of the output linear layer, one may absorb the nonlinearity into the loss function such that the model would still assume the aforementioned representation.

Considering the ending nonlinearity of \(G_1\) as the beginning component of the output layer, we can view the output layer as the KM output module and apply Theorem~\ref{th1}.
To be specific, write \(G_1 = \Phi\circ F_1\) for some \(F_1\), we can rewrite the model such that it satisfies the condition of Theorem~\ref{th1}:
\begin{align}
    \left(F(\mathbf{x})\right)_i = \langle \mathbf{w}_i, G_1(\mathbf{x}) \rangle + b_i =\langle \mathbf{w}_i, \Phi\left(F_1(\mathbf{x})\right) \rangle + b_i,\\
    k(F_1(\mathbf{u}), F_1(\mathbf{v})) := \langle \Phi(F_1(\mathbf{u})), \Phi(F_1(\mathbf{v})) \rangle.
\end{align}

Note that the assumption that \(\|\Phi(\mathbf{u})\|\) is fixed for all \(\mathbf{u}\) may require that one normalizes the activation vector/matrix/tensor in practice, which is the only modification one has to make for a standard NN to satisfy the conditions of the theorem.

\subsubsection{Loss Functions}
In terms of loss functions, many common ones including softmax \(+\) cross-entropy (two-class version), any monotonic nonlinearity \(+\) mean squared error, and hinge loss, admit the required representation.
We provide details in the Appendix.

\subsection{From Theory to Algorithm}
\label{algorithm}

To convert the theoretical result into an implementable learning algorithm for classification (with potentially more than two classes), one may proceed as follows.
Let an NN \(G_2\circ G_1\) be given, where \(G_2\) is a linear layer (absorb the trailing nonlinearity into the loss function if necessary) and \(G_1 = \Phi\circ F_1\) is a composition of arbitrary layers followed by nonlinearity \(\Phi\).
Suppose the activation vector of \(G_1\) is normalized such that it is a unit vector by (elementwise) dividing the activation vector by its norm.
Also let a loss function \(L\) be given and suppose it (or its two-class analog) satisfies the requirements of Theorem~\ref{th1}.
Define kernel \(k(F_1(\mathbf{u}), F_1(\mathbf{v})) = \langle \Phi(F_1(\mathbf{u})), \Phi(F_1(\mathbf{v})) \rangle\).
Determine \(\beta := \min k\) based on \(\Phi\).
Some examples include: \(\beta = 0\) for ReLU and sigmoid; \(\beta = -1\) for tanh.
Denote \(G_2\circ\Phi\) as \(F_2\).
The training consists of two stages: Training \(F_1\) and training \(F_2\) (without touching \(F_1\)).

{\bf Training \(F_1\).}
Given a batch of training data \(\{(\mathbf{x}_i, y_i)\}_{i=1}^n\) and let \(\mathcal{N}\) (for negative) denote all pairs of indices \(i, j\) such that \(y_i\neq y_j\) and \(\mathcal{P}\) (for positive) denote all pairs of indices \(i \neq j\) with \(y_i = y_j\).
Train \(F_1\) to \textit{maximize} one of the following proxy objective functions.
\begin{itemize}
    \item Alignment (negative only) (AL-NEO):
    \begin{equation}
        L_1(F_1) = \frac{\beta\sum_{(i, j)\in\mathcal{N}} k\left(F_1(\mathbf{x}_i), F_1(\mathbf{x}_j)\right)}{|\beta||\mathcal{N}|\sqrt{\sum_{(i, j)\in\mathcal{N}} \left(k\left(F_1(\mathbf{x}_i), F_1(\mathbf{x}_j)\right)\right)^2}};
    \end{equation}
    \item Contrastive (negative only) (CTS-NEO):
    \begin{equation}
    L_1(F_1) = -\frac{1}{|\mathcal{N}|}\sum_{(i, j)\in\mathcal{N}} \exp \left(k\left(F_1(\mathbf{x}_i), F_1(\mathbf{x}_j)\right)\right);
    \end{equation}
    \item Negative Mean Squared Error (negative only) (NMSE-NEO):
    \begin{equation}
        L_1(F_1) = -\frac{1}{|\mathcal{N}|}\sum_{(i, j)\in\mathcal{N}} \left(k\left(F_1({x}_i), F_1(\mathbf{x}_j)\right) - \beta\right)^2.
    \end{equation}
\end{itemize}
All of the above proxy objectives can be shown to learn an \(F_1\) that satisfies the optimality condition required by Theorem~\ref{th1}.

Note that in cases where some of these proxies are undefined (for example, when \(\beta = 0\)), we may train \(F_1\) to \textit{maximize} the following alternative proxy objectives instead assuming \(\alpha := \sup k\) is known, where we define \(k^\star_{ij}\) to be \(\alpha\) if \((i, j)\in\mathcal{P}\) or if \(i = j\) and \(\beta\) if otherwise.
Compared to the previous proxies, these are applicable to all \(k\)'s and impose a stronger constraint on learning.
Specifically, in addition to controlling the inter-class pairs, these proxies force intra-class pairs to share representations.
\begin{itemize}
    \item Alignment (AL): The kernel alignment~\cite{cristianini2002kernel} between the kernel matrix formed by \(k\) and \(k^\star\) on the given data.
    \item Upper Triangle Alignment (UTAL): Same as AL, except only the upper triangles minus the main diagonals of the two matrices are considered. This can be considered a refined version of the raw alignment.
    \item Contrastive (CTS):
    \begin{equation}
        L_1(F_1) = \frac{\sum_{(i, j)\in\mathcal{P}} \exp\left( k\left( F_1(\mathbf{x}_i), F_1(\mathbf{x}_j) \right) \right)}{\sum_{(i, j)\in\mathcal{N}\cup\mathcal{P}} \exp\left( k\left( F_1(\mathbf{x}_i), F_1(\mathbf{x}_j) \right) \right)};
    \end{equation}
    \item Negative Mean Squared Error (NMSE):
    \begin{equation}
        L_1(F_1) = -\frac{1}{n^2}\sum_{(i, j)}\left(k\left( F_1(\mathbf{x}_i), F_1(\mathbf{x}_j) \right) - k^\star_{ij}\right)^2.
    \end{equation}
\end{itemize}
Empirically, we found that alignment and squared error typically produced better results than the contrastive ones, which is potentially due to how the exponential term involved in the contrastive objectives made the range of ideal learning rates to be smaller.

\begin{figure*}[t]
    \centering
    \subfloat{
    \includegraphics[width=2\columnwidth]{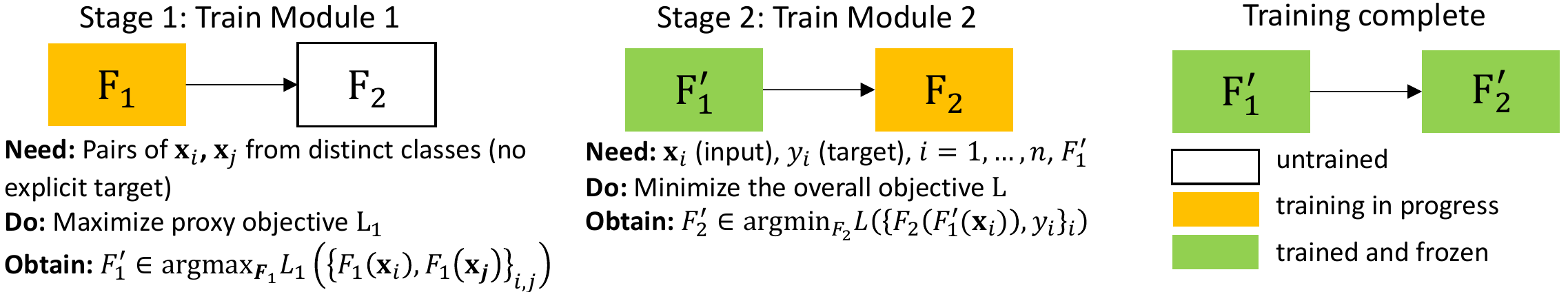}
    }
    \caption{
    The proposed modular training algorithm (two-module case) consists of two training stages.
    Suppose we are given a two-module model \(F_2\circ F_1\) and an overall classification objective function \(L(F_2\circ F_1)\), e.g., cross-entropy with weight regularization.
    First, the input module \(F_1\) is trained to maximize a proxy hidden objective as defined in Sec.~\ref{algorithm}.
    Then this input module is frozen at, say, \(F_1^\prime\).
    And the output module is trained to minimize \(L(F_2\circ F_1^\prime)\).
    Note that during the training of the input module does not involve training the output module, and vice versa.
    In other words, the training process is fully modular.
    }
    \label{fig10}
\end{figure*}

{\bf Training \(F_2\).}
Now suppose \(F_1\) has been trained and frozen at \(F_1^\prime\), we simply train \(F_2\) to minimize the overall loss function \(L(F_2\circ F_1^\prime)\).
This process is illustrated in Fig.~\ref{fig10}.

\section{A Method for Task Transferability Estimation}
\label{sec3}

In this section, we demonstrate the improved module reusability of modularized DL.
Let a source task be defined as one that we have already solved and a target task be one that we are interested in solving.
Specifically, we consider the following practical issue in transfer learning:
Assume one is given a set of input modules pre-trained on some source tasks and a target task to be solved by training an output module on top of a frozen input module.
Which input module will result in the most performant network on the target task?

Underlying this practical problem is a theoretical issue that is central to many important research domains including transfer learning, continual/lifelong learning, meta learning, and multi-task learning:
Given a set of tasks, how to effectively model the task space structure in terms of the transferability between task pairs, i.e., how helpful knowledge about a source task is for solving a target task~\cite{zamir2018taskonomy,achille2019task2vec,tran2019transferability,nguyen2020leep}.

We propose the following solution to the module reusability (and also task transferability) problem.
Let a target task be characterized by the loss function \(L(F_2\circ F_1, S)\), where \(L\) is a loss function assuming the representation in Theorem~\ref{th1} and \(S\) is some training data~\cite{tran2019transferability}.
Our modular learning framework suggests that we can measure the goodness of these pre-trained \(F_1\)'s for this target task by measuring \(L_1(F_1, S)\), where \(L_1\) is a proxy objective.
This is based on the fact that the \(F_1\) that maximizes the proxy objective \(L_1(F_1, S)\) constitutes the input module of a minimizer for \(L(F_2\circ F_1, S)\).
Therefore, one can simply rank the pre-trained \(F_1\)'s in terms of how well they maximize \(L_1(F_1, S)\) and select the maximizer.
Since a well-trained input module must encode useful information for the source task, this procedure can also fully describe the task space structure of the given tasks.
The benefit is that this procedure requires no training.
Indeed, one only needs to run the given modules on potentially a subset of \(S\).
Moreover, this method is task agnostic, flexible, and completely data-dependent.

In terms of the optimality of our transferability measure, if we define the true transferability of a particular \(F_1\) to be \(\min_{F_2}\mathbb{E}_{(\mathbf{x}, y)}L(F_2\circ F_1, (\mathbf{x}, y))\)~\cite{tran2019transferability}, then \(\min_{F_2}L(F_2\circ F_1, S)\) is a bound on the true transferability minus a complexity measure on the model class from which we choose our model~\cite{bartlett2002rademacher}.
We leave a more refined study as future work.

Comparisons of our method against some notable related work are provided in Sec.~\ref{transferability related} and summarized in Table~\ref{table3}, from which it is clear that our method is among the fastest and most flexible.

\section{Related Work}
\subsection{Connecting Neural Networks With Kernel Methods}
While some works establish connections via exactly matching one architecture to the other, others do so from a probabilistic perspective by studying large-sample behavior in the infinite layer widths limit and/or taking the expectation over the network parameters.
The former line of research, to which this work belongs, often yields more direct and practical results since the theories operate under much milder assumptions.

\subsubsection{Exact Equivalences}
In \cite{vapnik2000}, a family of kernels were defined to mimic single-hidden-layer MLPs.
The resulting KMs bear the same mathematical formulations as the corresponding NNs with the constraint that the input layer weights of the NNs are fixed.
The authors of \cite{suykens1999training} modified these kernels to allow the KMs to correspond to fully-trainable single-hidden-layer MLPs.
Their construction can be viewed as a special case of ours.
Nevertheless, they presented their work as another way to train MLPs without pointing out the connections with KMs, i.e., their MLPs are in fact also KMs.
The input and output layers are trained alternately, with the former learning to minimize the VC dimension~\cite{vapnik1971uniform} of the latter while the latter learning to classify.
An optimality guarantee of the training was hinted.
In contrast, we extended their theoretical framework, explicitly established the connections between NNs and KMs, and proposed a fully modular training approach with proved strong optimality guarantee.

\subsubsection{Equivalences in Infinite Widths and/or in Expectation}
That single-hidden-layer MLPs are Gaussian processes in the infinite width limit and in expectation of random input layer has been known at least since~\cite{neal1995bayesian}.
\cite{lee2017deep} generalized the classic result to deeper MLPs.
\cite{cho2009kernel} defined a family of ``arc-cosine'' kernels to imitate the computations performed by infinitely wide networks in expectation.
\cite{shankar2020neural} proposed kernels that are equivalent to expectations of finite-widths random networks.
\cite{arora2019exact} presented exact computations of some kernels, using which the kernel regression models can be shown to be the limit (in widths and training time) of fully-trainable, infinitely wide fully-connected networks trained with gradient descent.
In comparison, our work established that NNs can be directly viewed as KMs, requiring neither infinite widths nor taking expectation over random parameters.


\subsection{Modularized Deep Learning}

NNs were developed with inspiration from human brain structures and functions~\cite{kandel2000principles}.
The human brain itself is modular in a hierarchical manner, as the learning process always occurs in a very localized subset of highly inter-connected nodes which are relatively sparsely connected to nodes in other modules~\cite{hrycej1992modular,meunier2010modular}.
That is to say that the human brain is organized as functional, sparsely connected subunits~\cite{clune2013evolutionary}.


Many existing works in machine learning can be analyzed from the perspective of modularization.
An old example is the mixture of experts~\cite{jacobs1991adaptive,jordan1994hierarchical} which uses a gating function to enforce each expert in a committee of networks to solve a distinct group of training cases.
Another recent example is the generative adversarial networks (GANs)~\cite{goodfellow2014generative}.
Typical GANs have two competing neural networks that are essentially decoupled in functionality and can be viewed as two modules.
In~\cite{watanabe2018modular}, the authors proposed an \textit{a posteriori} method that analyzes a trained network as modules in order to extract useful information.
Most works in this direction, however, achieved only partial modularization due to their dependence of end-to-end optimization.

The authors of \cite{fahlman1990cascade} pioneered the idea of greedily learn the architecture of an NN, achieving full modularization.
In their work, each new node is added to maximize the correlation between its output and the residual error.
Several works attempted to remove the need for end-to-end backpropagation via approximating gradient signals locally at each layer or each node \cite{bengio2014auto,lee2015difference,carreira2014distributed,balduzzi2015kickback,jaderberg2017decoupled}.
In~\cite{zhou2017deep}, a backpropagation-free deep architecture based on decision trees was proposed.
In~\cite{lowe2019putting}, the authors proposed to learn the hidden layers with unsupervised contrastive learning, decoupling their training from that of the output layer.
Compared to these existing methods that enable full modularization, our method is simple to implement and provide strong optimality guarantee.
A similar provably optimal modular training framework was proposed in~\cite{duan2019kernel}.
However, their optimality result was less general than ours and their framework required that the NN be modified by substituting certain neurons with KMs using classical kernels that have quadratic runtime.

\subsection{Task Transferability}
\label{transferability related}
Describing the task space structure via measuring the transferability between tasks, i.e., estimating to what extent representations learned from one task can help learning another, is a central problem in transfer learning, multi-task learning, meta learning, continual/lifelong learning, etc~\cite{zamir2018taskonomy,achille2019task2vec,tran2019transferability,nguyen2020leep}.

Information-theoretic measures have been proven useful in quantifying task transferability.
Early task-relatedness measures include the \(\mathcal{F}\)-relatedness~\cite{ben2003exploiting} and the \(\mathcal{A}\)-distance~\cite{ben2007analysis}.
\(\mathcal{H}\)-divergence~\cite{ganin2016domain} and Wasserstein distance~\cite{li2018extracting} have received increasing attention in recent years.
Specifically, \cite{liu2017adversarial} applied \(\mathcal{H}\)-divergence in natural language processing for text classification, whereas~\cite{janati2018wasserstein} used Wasserstein distance to estimate the similarity of linear parameters instead of the data generation distributions.
Along this line of research, some more recent methods include H-score~\cite{bao2019information} and the Bregman-correntropy conditional divergence~\cite{yu2020measuring}, the latter of which used the correntropy functional~\cite{liu2007correntropy} and the Bregman matrix divergence~\cite{kulis2009low} to quantify divergence between mappings.

Model-based methods, i.e., methods that leverage trained models to extract knowledge about tasks, are more closely related to our proposed method.
\cite{tran2019transferability} estimated task transferability via the negative conditional entropy between their training label sequences, assuming the two tasks share the same input data examples.
The proposed method assumed the existence of an optimal trained source model but did not actually use a trained model in the estimation.
\cite{nguyen2020leep} removed the input-sharing assumption.
Both works assumed that the source and the target tasks use the cross-entropy loss.
Our method works, on the other hand, asserts mild assumptions on the choice of loss function only.
Taskonomy~\cite{zamir2018taskonomy} estimated the transferability of a pair of tasks via the transfer performance from the tasks to a third reference task.
Task2Vec~\cite{achille2019task2vec} tuned a single ``probe'' network on all target tasks and extracted task-characterizing information via the parameters of this network.
Both Taskonomy and Task2Vec required training models on target tasks to be able to extract task information.
In contrast, our method does not require any training on target tasks.

All of the mentioned model-based methods have their limiting assumptions.
The main assumption of our method, required by our optimality guarantee, is that the output module (classifier) that will be tuned on top of the transferred component admits a particular form as described in Theorem~\ref{th1}.
Essentially, it is required to be a single-layer NN.
It is worth noting that this set-up is common in practice.
We summarize the properties of these related methods in Table~\ref{table3}.
\begin{table}
    \centering
    \begin{tabular}{lllll}
        \hline
        Method & Training & Test Runtime & Main Assumption \\
        \hline
        TASK2VEC~\cite{achille2019task2vec} & \cmark & & a reference model \\
        Taskonomy~\cite{zamir2018taskonomy} & \cmark & & a third reference task \\
        NCE~\cite{tran2019transferability} & \xmark & \(\mathcal{O}(n^3)\) & tasks share input \\
        LEEP~\cite{nguyen2020leep} & \xmark & \(\mathcal{O}(n^2)\) & cross-entropy loss \\
        Ours & \xmark & \(\mathcal{O}(n^2)\) & form of classifier \\
        \hline
    \end{tabular}
    \caption{
        Comparisons with similar methods for task transferability estimation.
        For the training-free methods, we also provide the test runtime in terms of \(n\), the size of the target task dataset that is being used for the estimation.
        Note that the extra trainings constitute the main overhead of the methods that require training, making comparisons on their test runtime against the training-free methods meaningless.
        Details are provided in Sec.~\ref{transferability related}.
    }
    \label{table3}
\end{table}

\section{Experiments}

\subsection{Sanity Check: Modular Training Results in Identical Learning Dynamics As End-to-End}
In this section, we attempt to answer the important question: Do end-to-end and the proposed modular training, when restricted to the architecturally identical hidden modules, drive the underlying modules to functions that are the same in terms of minimizing the overall loss function?
Clearly, a positive answer would verify empirically the optimality of the modular approach.

We now test this hypothesis with toy data.
The set up is as follows.
We generate 1000 32-dimensional random input and assign them random labels from 10 classes.
The underlying network consists two modules.
The input module is a \((32 \to 512)\) fully-connected (fc) layer followed by ReLU nonlinearity and another \((512 \to 2)\) fc layer.
The output module is a \((2 \to 10)\) fc layer.
The two modules are linked by a tanh nonlinearity (output normalized to unit vector).
The overall loss is the cross-entropy loss.
And for modular training, the proxy objective is CTS-NEO.
We visualize the activations from the tanh nonlinearity as an indicator of the behavior of the hidden layers under training.

\begin{figure}[t]
    \centering
    \subfloat{
        \includegraphics[width=.3\columnwidth]{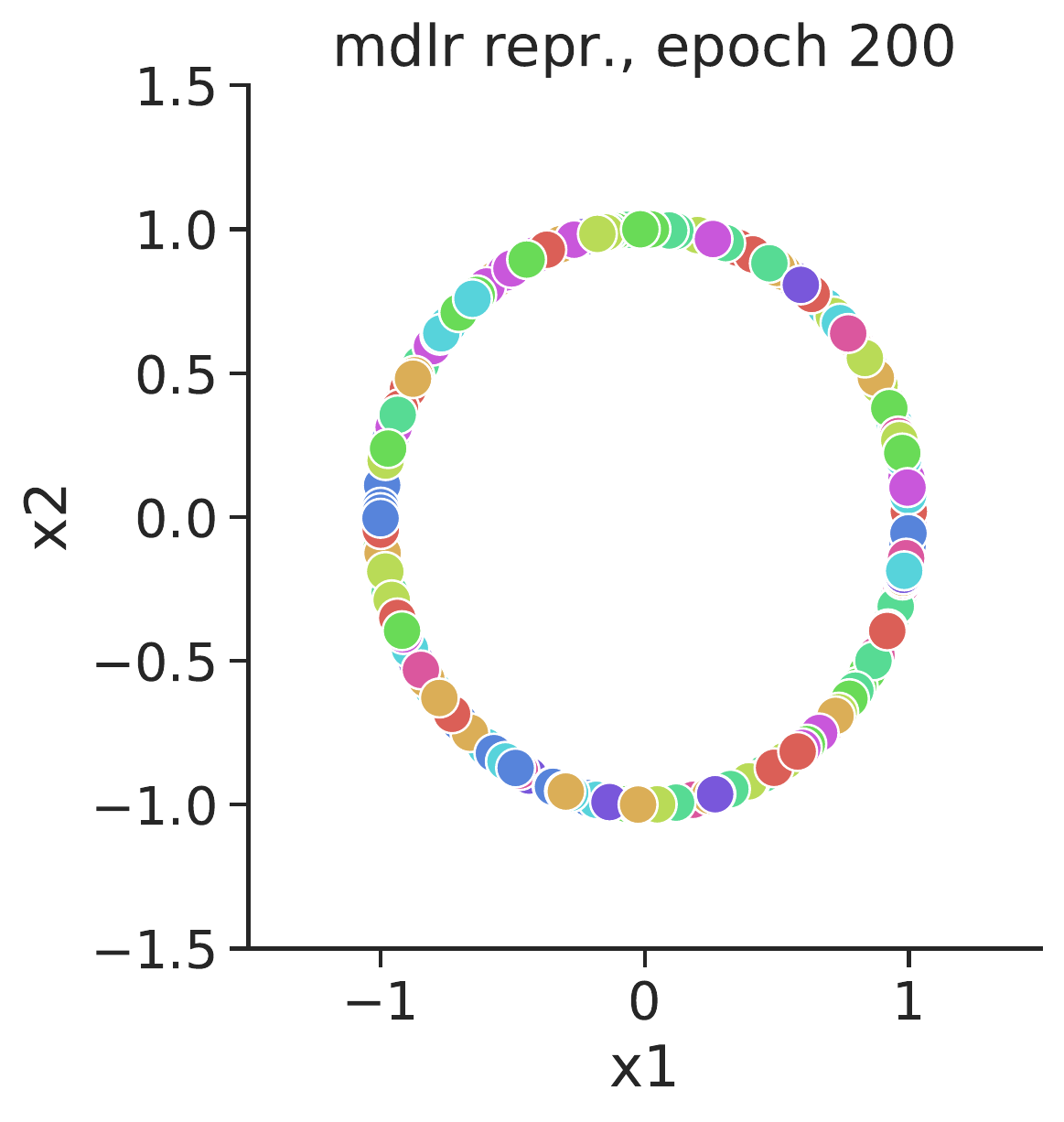}
    }
    \subfloat{
        \includegraphics[width=.3\columnwidth]{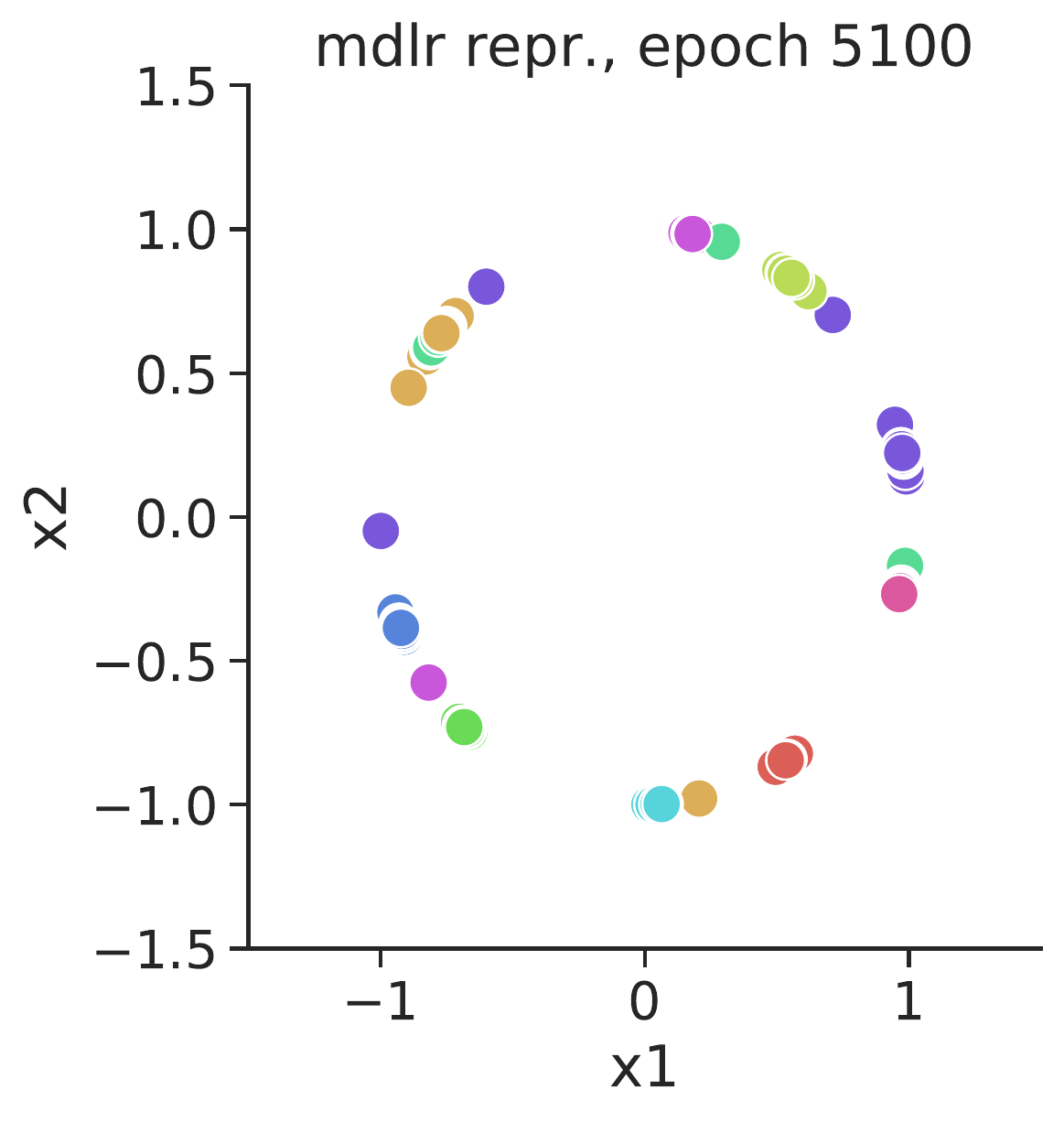}
    }
    \subfloat{
        \includegraphics[width=.3\columnwidth]{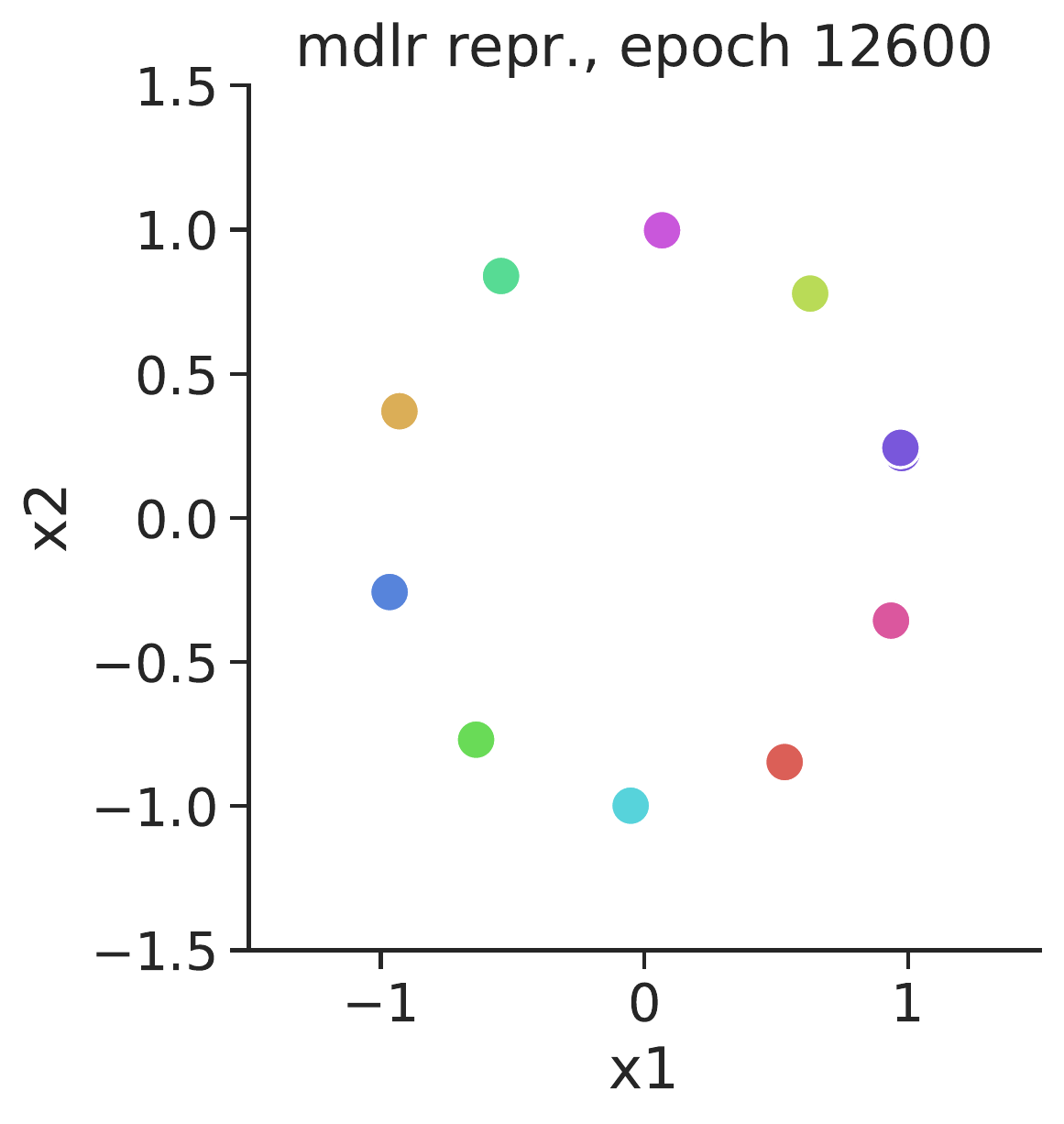}
    }
    \vspace{-10pt}
    \subfloat{
        \includegraphics[width=.3\columnwidth]{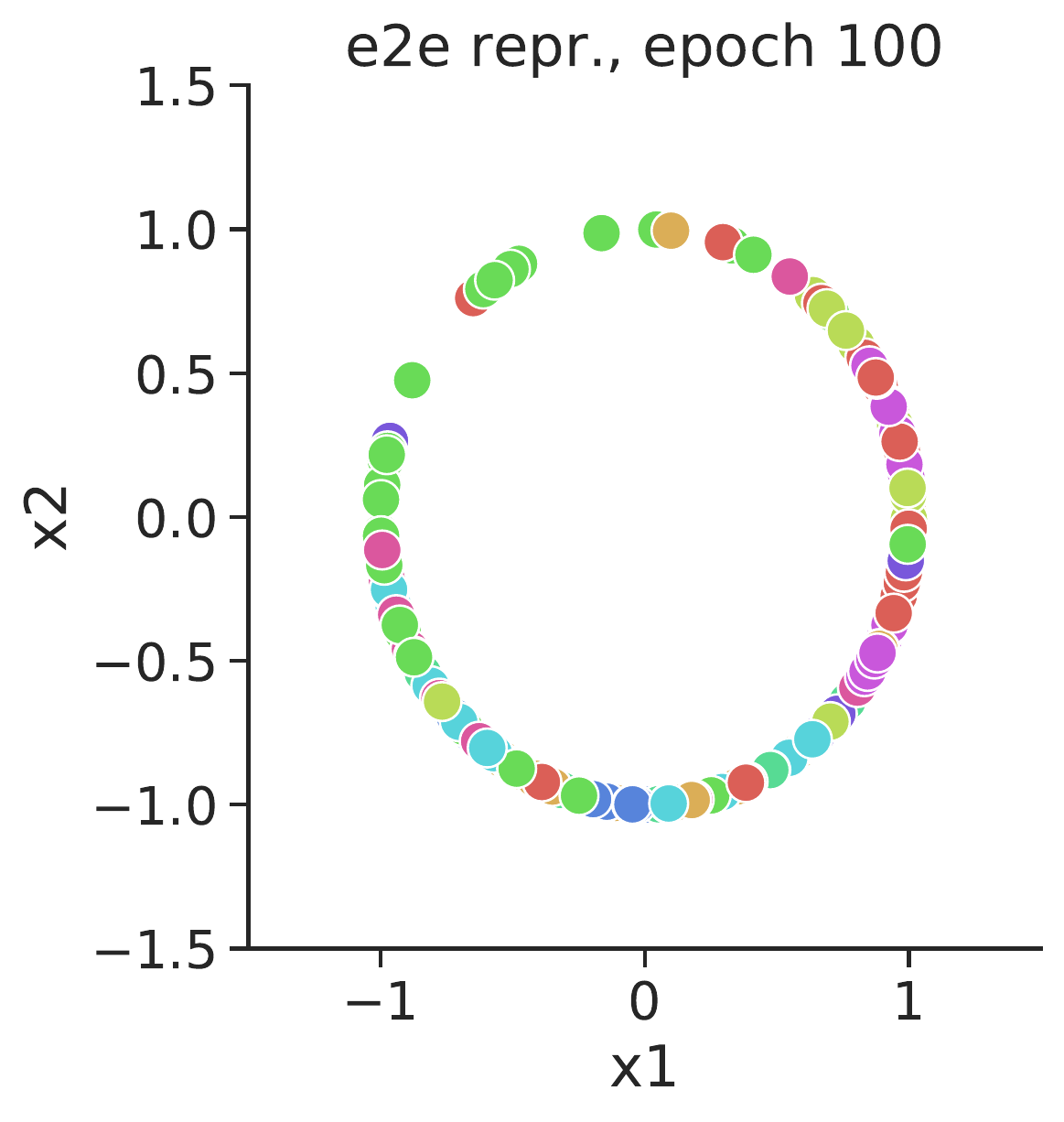}
    }
    \subfloat{
        \includegraphics[width=.3\columnwidth]{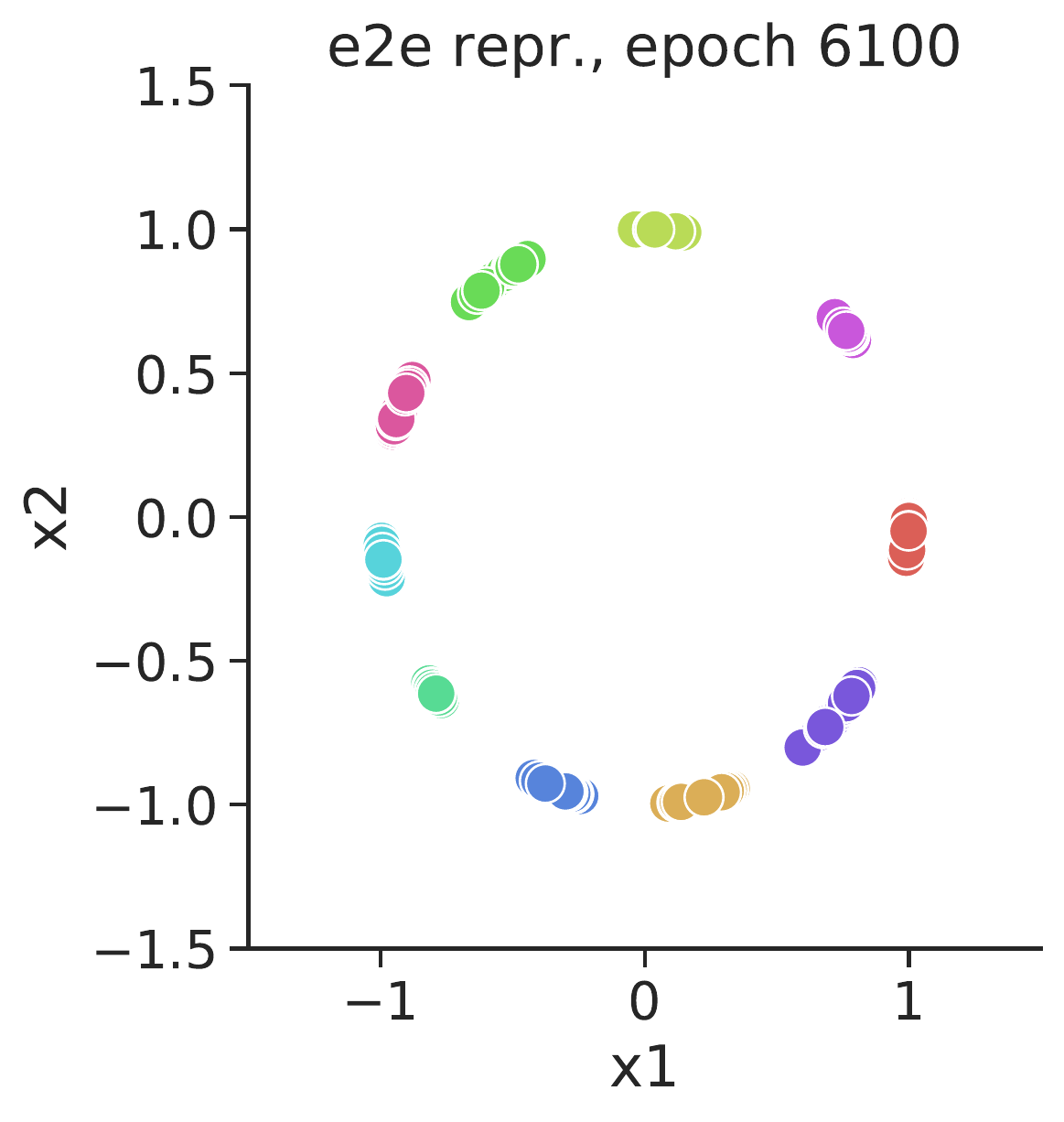}
    }
    \subfloat{
        \includegraphics[width=.3\columnwidth]{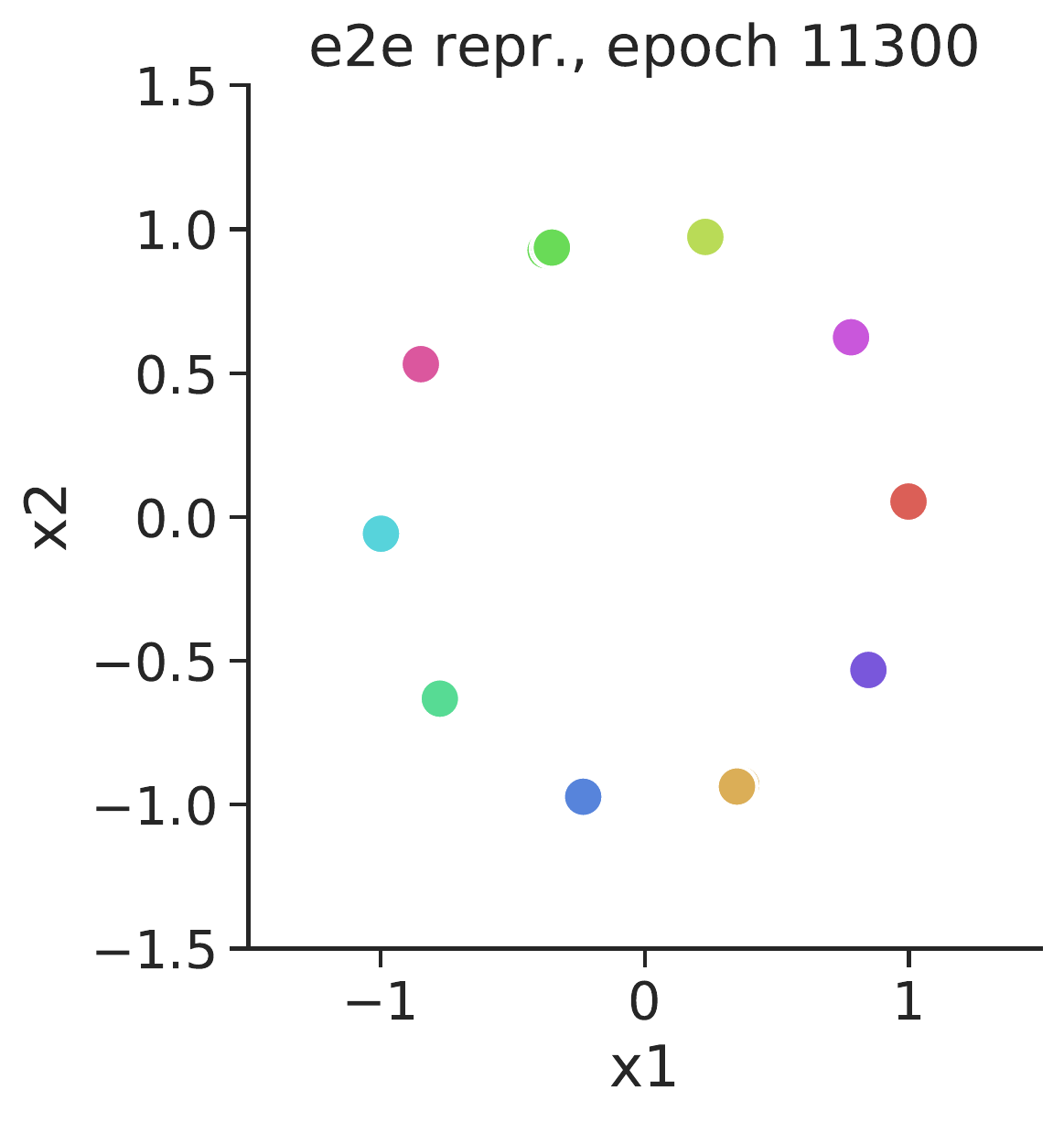}
    }
    \caption{
        Using toy data, we visualize the learning dynamics of our modular training approach and that of backpropagation by visualizing the output representations from the input module.
        We observe that the two methods result in input modules that are identical functions when restricted to the random training data, confirming that one may greedily optimize the input module using a proxy objective and still obtain the same outcome as an end-to-end approach.
        e2e for end-to-end training.
    mdlr for modular training.
    Classes are color-coded.
    }
    \label{fig3}
\end{figure}

From Fig.~\ref{fig3}, we see that the underlying input modules are indeed driven to the same functions (when restricted to the training data and in terms of minimizing the overall loss) by both modular and end-to-end in the limit of training time going to infinity.
This confirms the optimality of our proposed method (albeit in a simplified set-up) and allows us to modularize learning with confidence.

\subsection{Sanity Check: Proxy Objectives Align Well With Accuracy}
We now extend the verification of optimality from the simplified set-up in the previous section to a more practical setting.
Recall that we have characterized our proxy objective as function of the input module whose maximizers constitute parts of the overall loss minimizers and we have proposed to train the input module to maximize the proxy objective.

Ideally, however, a proxy objective for input module should satisfy: The overall accuracy is a function of solely the proxy value and the output module and as a function of the proxy value, the overall accuracy is strictly increasing.\footnote{This type of results is reminiscent of the ``ideal bounds'' for representation learning sought for in, e.g.,~\cite{arora2019theoretical}.}
We now demonstrate empirically that the proxies we proposed enjoy this property.

The set-up is as follows.
We train a LeNet-5 as two modules on MNIST with \((\text{conv1} \to \text{tanh} \to \text{max pool} \to \text{conv2} \to \text{tanh} \to \text{max pool} \to \text{fc1} \to \text{tanh} \to \text{fc2})\) as the input module, and \((\text{tanh} \to \text{fc3})\) as the output one.
The input module is trained with a number of different epochs so as to achieve different proxy values.
And for each epoch, we freeze the input module and train output module to minimize the overall cross-entropy until it converges.
Mathematically, suppose we froze input module at \(F_1^\prime\) and we denote the proxy as \(L_1\) and overall accuracy as \(A\), we are visualizing \(\max_{F_2}A(F_2\circ F_1^\prime)\) vs. \(L_1(F_1^\prime)\), where \(F_2\) is the output module.

\begin{figure}[!t]
    \centering
    \includegraphics[width=.8\columnwidth]{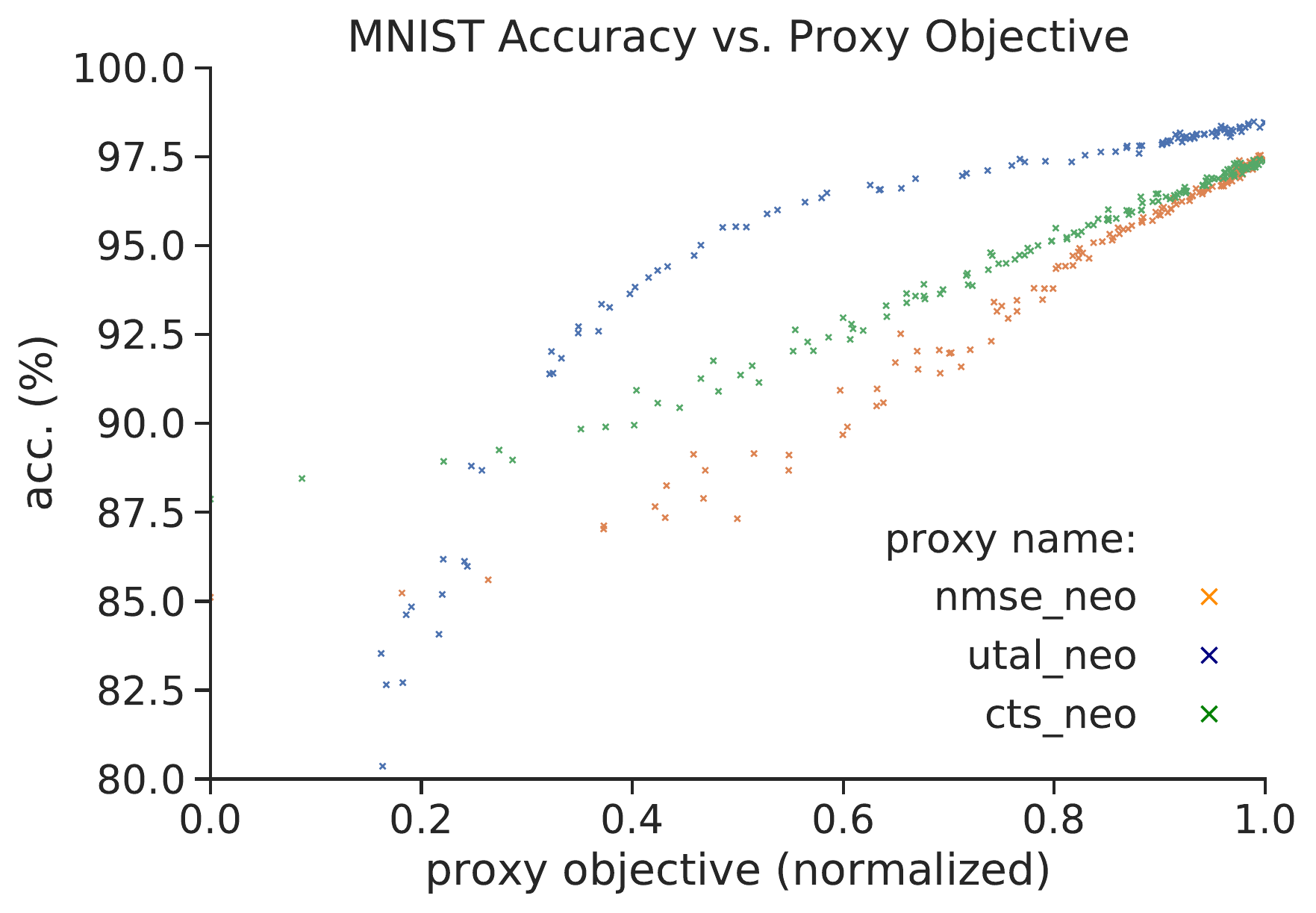}
    \caption{
        The overall accuracy, as a function of the proxy objective, is increasing.
        The positive correlation becomes near perfect in high-performance regime, validating our theoretical results.
        Overall, this justifies the optimality of training the input module to maximize the proxy objective.
        Note that the values of the illustrated proxies were normalized to \([0, 1]\) in order for them to be properly visualized in the same plot.
    }
    \label{fig4}
\end{figure}

Fig.~\ref{fig4} shows that the overall accuracy, as a function of the proxy value, is indeed approximately increasing.
Further, this positive correlation becomes near perfect in high-accuracy regime, which agrees with our theoretical results.
To summarize, we have extended our theoretical guarantees and empirically verified that maximizing the proposed proxy objectives effectively learns input modules that are optimal in terms of maximizing the overall accuracy, rendering end-to-end training unnecessary.

\subsection{Accuracy on MNIST and CIFAR-10}
We now present the main results on MNIST and CIFAR-10 demonstrating the effectiveness of our modular learning method.
To facilitate fair comparisons, end-to-end and modular training operate on the same backbone network.
For all results, we used stochastic gradient descent as the optimizer with batch size \(128\).
For each module in the modular method as well as the end-to-end baseline, we trained with annealing learning rates (\(0.1, 0.01, 0.001\), each for \(200\) epochs).
The momentum was set to \(0.9\) throughout.
For data preprocessing, we used simple mean subtraction followed by division by standard deviation.
On CIFAR-10, we used the standard data augmentation pipeline from~\cite{he2016deep}, i.e., random flipping and clipping with the exact parameters specified therein.
In modular training, the models were trained as two modules, with the output layer alone as the output module and everything else as the input module as specified in Sec.~\ref{two-module view}.
We normalized the activation vector right before the output layer of each model to a unit vector such that the equal-norm condition required by Theorem~\ref{th1} is satisfied.
We did not observe a significant performance difference after this normalization.

From Table~\ref{table1} and \ref{table2}, we see that on three different backbone networks and both MNIST and CIFAR-10, our modular learning compares favorably against end-to-end.
Since under two-module modular training, the ResNets can be viewed as KMs with adaptive kernels, we also compared performance with other NN-inspired kernel methods in the literature.
In Table~\ref{table2}, the ResNet KMs clearly outperformed other kernel methods by a considerable margin.

\begin{table}
    \centering
    \begin{tabular}{lllc}
        \hline
        Model & Training & Obj. Fn. & Acc. (\%)\\
        \hline
        LeNet-5 (ReLU) & e2e & XE & 99.33 \\
        LeNet-5 (tanh) & e2e & XE & 99.32 \\
        \hline
        LeNet-5 (ReLU) & mdlr & AL/XE & 99.35 \\
        LeNet-5 (ReLU) & mdlr & UTAL/XE & \textbf{99.42} \\
        LeNet-5 (ReLU) & mdlr & MSE/XE & 99.36 \\
        LeNet-5 (tanh) & mdlr & AL(-NEO)/XE & 99.11 (99.19) \\
        LeNet-5 (tanh) & mdlr & UTAL(-NEO)/XE & 99.21 (99.11) \\
        LeNet-5 (tanh) & mdlr & MSE(-NEO)/XE & 99.27 (99.23) \\
        LeNet-5 (tanh) & mdlr & CTS(-NEO)/XE & 99.16 (99.16) \\
        \hline
    \end{tabular}
    \caption{
        Accuracy on MNIST.
        e2e for end-to-end.
        mdlr for modular as two modules.
        Obj. Fn. specifies the loss function (and the proxy objective for the input module, if applicable) used.
        XE stands for cross-entropy.
        Definitions of the proxy objectives can be found in Sec.~\ref{algorithm}.
        Using LeNet-5 as the network backbone on MNIST, our modular approach compares favorably against end-to-end training.
    }
    \label{table1}
\end{table}

\begin{table}
    \centering
    \begin{tabular}{llllc}
        \hline
        Model & Data Aug. & Training & Obj. Fn. & Acc. (\%)\\
        \hline
        ResNet-18 & flip \& clip & e2e & XE & 94.91 \\
        ResNet-152 & flip \& clip & e2e & XE & \textbf{95.87} \\
        \hline
        ResNet-18 & flip \& clip & mdlr & AL/XE & 94.93 \\
        ResNet-152 & flip \& clip & mdlr & AL/XE & 95.73 \\
        \hline
        CKN~\cite{mairal2014convolutional} & none & & & 82.18 \\
        CNTK~\cite{li2019enhanced} & flip & & & 81.40 \\
        CNTK\(^*\)~\cite{li2019enhanced} & flip & & & 88.36 \\
        CNN-GP~\cite{li2019enhanced} & flip & & & 82.20 \\
        CNN-GP\(^*\)~\cite{li2019enhanced} & flip & & & 88.92 \\
        NKWT\(^*\)~\cite{shankar2020neural} & flip & & & 89.80 \\
        \hline
    \end{tabular}
    \caption{
        Accuracy on CIFAR-10.
        Modular training yielded favorable results compared to end-to-end.
        To the best of our knowledge, there is no other purely modular training method that matches backpropagation with a competitive network backbone on CIFAR-10~\cite{jaderberg2017decoupled,lee2015difference,carreira2014distributed,lowe2019putting,duan2019kernel}.
        The ResNets trained with our modular approach can be viewed as kernel machines and outperformed other existing NN-inspired kernel methods.
        \(*\) means the method used more sophisticated data preprocessing than the ResNets.
        Note that all of the baseline kernel methods have quadratic runtime and it is nontrivial to incorporate data augmentation into their pipelines.
    } 
    \label{table2}
\end{table}


\subsection{Label Efficiency of Modular Deep Learning}
\label{sec4}
Our modular training needs only implicit labels (weak supervision) for training the latent modules.
Full supervision is only needed for the output module.
And we argue that in fact, the modular training of output module requires a smaller number of fully-labeled data than end-to-end training.

Intuitively, the input module in a deep architecture can be understood as learning a new representation of the given data with which the output module's classification task is simplified.
A simple way to quantify the difficulty of a learning task is through its sample complexity, which, when put in simple terms, refers to the number of labeled examples needed for a given model and a training algorithm to achieve a certain level of test accuracy~\cite{shalev2014understanding}.

In a modular training setting, one can decouple the training of input and output modules and then it would make sense to discuss the sample complexity of the two modules individually.
The output modules should require fewer labeled data examples to train since its task has been simplified when the input one has been well-trained.
We now observe that this is indeed the case.

\begin{figure}[!t]
    \centering
    \includegraphics[width=1\columnwidth]{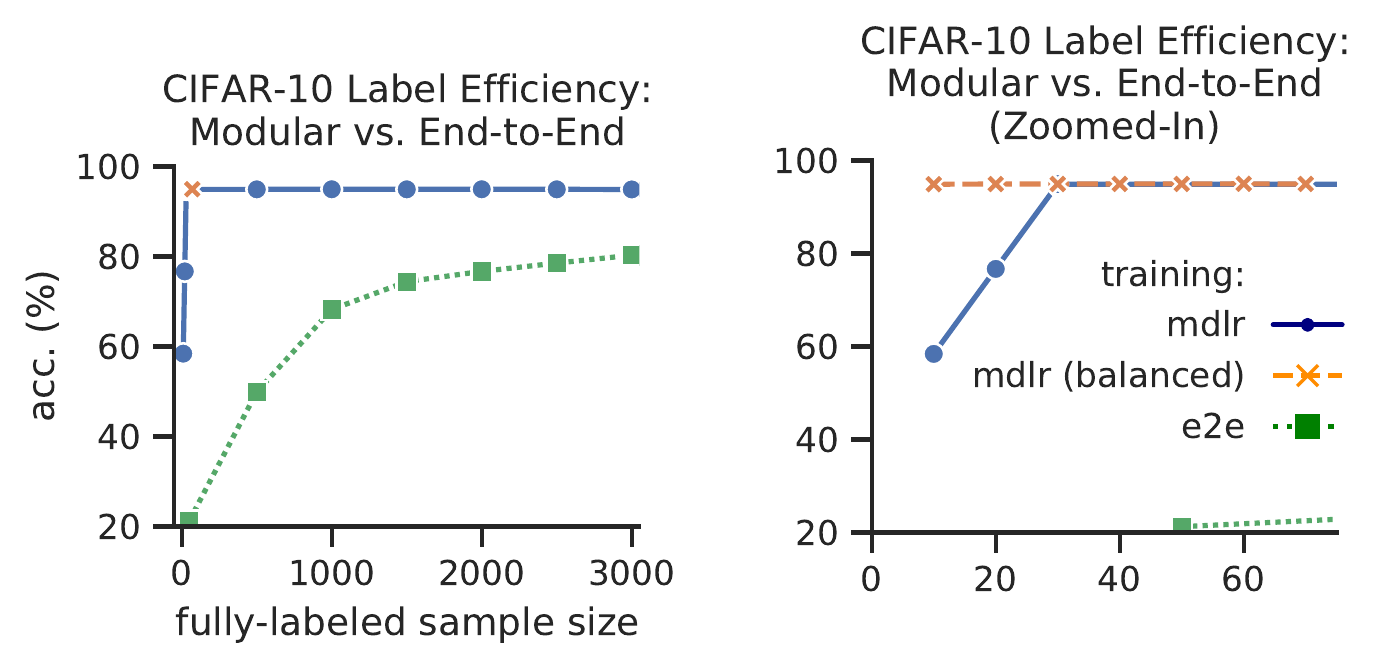}
    \caption{
        With the output module trained with only \(30\) randomly chosen fully-labeled examples, the modular model still achieved \(94.88\%\) accuracy on CIFAR-10 (the same model achieved \(94.93\%\) when using all \(50000\) labeled examples).
        When the training data has balanced classes (mdlr (balanced)), modular training only required \(10\) randomly chosen examples to achieve \(94.88\%\) accuracy --- a single example per class.
    }
    \label{fig5}
\end{figure}

We trained ResNet-18 on CIFAR-10 with end-to-end and our modular approach.
The input module was trained with the full training set in the modular method, but again, this only requires implicit pairwise labels.
We now compare the need for fully-labeled data between the two training methods.

With the full training set of \(50000\) labeled examples, the modular and end-to-end model achieved \(94.93\%\) and \(94.91\%\) test accuracy, respectively.
From Fig.~\ref{fig5}, we see that while the end-to-end model struggled in the label-scarce regime, achieving barely over \(20\%\) accuracy with \(50\) fully-labeled examples, the modular model consistently achieved strong performance, achieving \(94.88\%\) accuracy with \(30\) \textit{randomly chosen} fully-labeled examples for training.
In fact, if we ensure that there is at least one example from each class in the training data, the modular approach needed as few as \(10\) \textit{randomly chosen} examples to achieve \(94.88\%\) accuracy, that is, \textit{a single randomly chosen example per class}.\footnote{The fact that the modular model underperformed at \(10\) and \(20\) labels is likely caused by that there were some classes missing in the training data that we randomly selected. Specifically, \(4\) classes were missing in the \(10\)-example training data, and \(2\) in the \(20\)-example data.}

These observations suggest that our modular training method for classification can almost completely rely on implicit pairwise labels and still produce highly performant models, which suggests new paradigms for obtaining labeled data that can potentially be less costly than the existing ones.
This also indicates that the current form of strong supervision, i.e., full labels on individual data examples, relied on by existing end-to-end training, is not efficient enough.
In particular, we have shown that by leveraging pairwise information in the RKHS, implicit labels in the form of whether pairs of examples are from the same class are just as informative.
This may have further implications for un/semi-supervised learning.

\subsubsection{Connections With Un/Semi-Supervised Learning}

Whenever the training batch size is less than the size of the entire training set, we are using \textit{strictly less} supervision than using full labels.\footnote{This is because if one does not cache information over batches, it is impossible to recover the labels for the entire training set if one is only given pairwise relationships on each batch.}
The fact that our learning method still produces equally performant models suggests that we are using supervision more efficiently.

Existing semi-supervised methods also aim at utilizing supervision more efficiently.
They still rely on end-to-end training and use full labels, but achieve enhanced efficiency mostly through data augmentation.
While our results are not directly comparable to those from existing methods (these methods do not work in the same setting as ours: They use \textit{fewer} full labels while we use mostly \textit{implicit} labels), as a reference point, the state-of-the-art semi-supervised algorithm achieves \(88.61\%\) accuracy on CIFAR-10 with \(40\) labels with a stronger backbone network (Wide ResNet-28-2) and a significantly more complicated training pipeline~\cite{sohn2020fixmatch}.

Therefore, we think the insights gained from our modular learning suggest that (1) the current form of supervision, i.e., full labels on individual data examples, although pervasively used by supervised and semi-supervised learning methods, is far from efficient and (2) by leveraging novel forms of labels and learning methods, simpler yet stronger un/semi-supervised learning algorithms exist.
We leave this as a future work.

\subsection{Transferability Estimation With Proxy Objective}

To empirically verify the effectiveness of our transferability estimation method, we created a set of tasks by selecting \(6\) classes from CIFAR-10 and grouping each pair into a single binary classification task.
The classes selected are: \textit{cat, automobile, dog, horse, truck, deer}.
Each new task is named by concatenating the first two letters from the classes involved, e.g., \textit{cado} refers to the task of classifying dogs from cats.

We trained one ResNet-18 using the proposed modular method for each source task, using AL as the proxy objective.
For each model, the output linear layer plus the nonlinearity preceding it is the output module, and everything else belongs to the input module.
All networks achieved on average \(98.7\%\) test accuracy on the task they were trained on\footnote{Highest accuracy was \(99.95\%\), achieved on \textit{audo}. Lowest was \(93.05\%\), on \textit{cado}.}, suggesting that each frozen input module possesses information essential to the source task.
Therefore, the question of quantifying input module reusability is essentially the same as describing the transferability between each pair of tasks.

The true transferability between a source and a target task in the task space can be quantified by the test performance of the optimal output module trained for the target task on top of a frozen input module from a source task~\cite{tran2019transferability}.
Our estimation of transferability is based purely on proxy objective: A frozen input module achieving higher proxy objective value on a target task means that the source and target are more transferable.
This estimation is performed using a randomly selected subset of the target task training data.

\begin{figure}[t]
    \centering
    \subfloat{
        \includegraphics[width=\columnwidth]{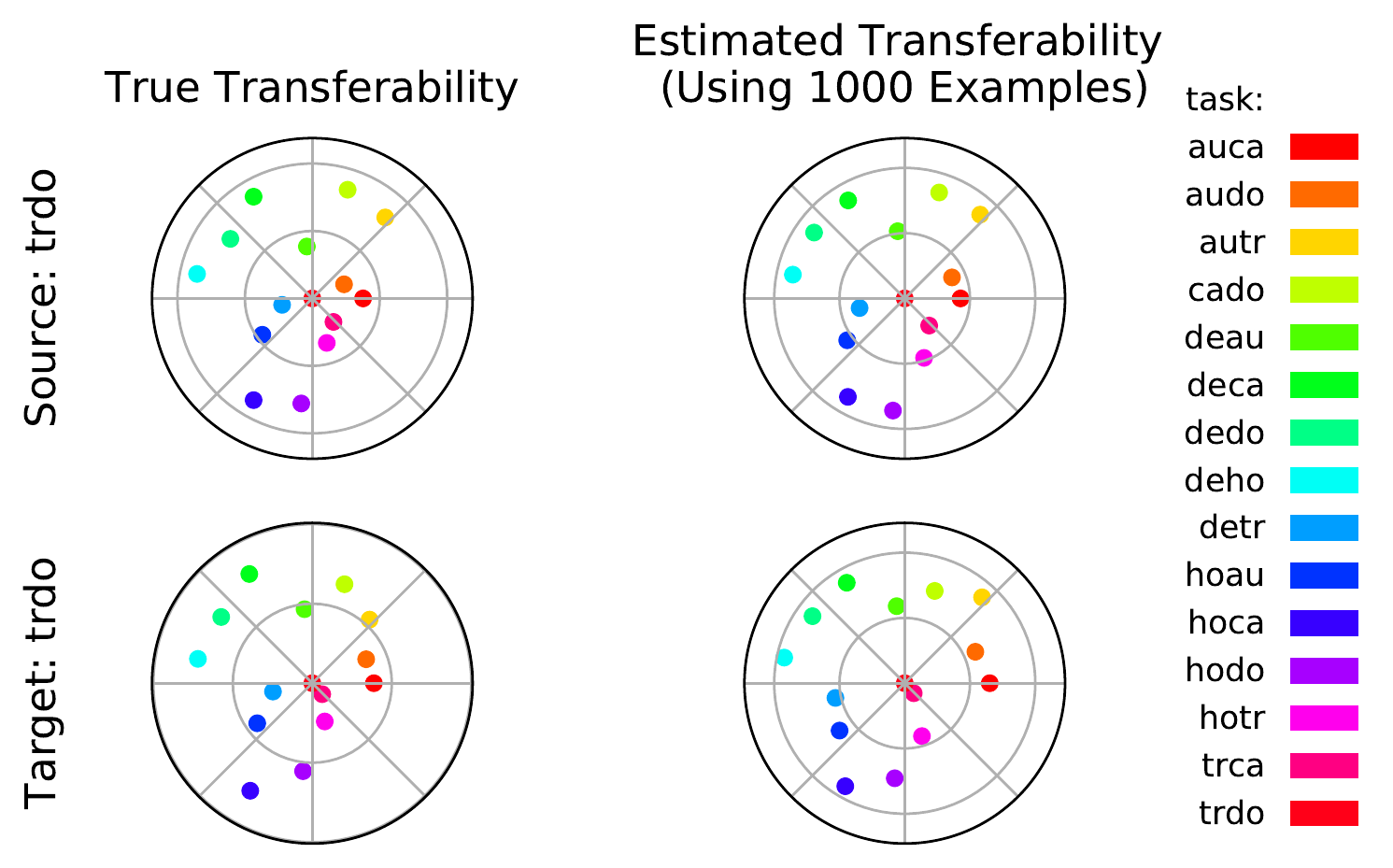}
    }
    \caption{
    Each task is assigned a distinct color and an angle in polar coordinate.
    Transferability with respect to \textit{trdo} is illustrated using the distance from the origin, with a smaller distance indicating a higher transferability.
    At practically no computational overhead, our proposed method correctly described the task space structure using only \(10\%\) randomly selected training data from target task.
        Plots with other \(14\) tasks being the source and target are provided in the Appendix.
    }
    \label{fig6}
\end{figure}

In Fig.~\ref{fig6}, we visualize the target space structure with \textit{trdo} being the source and the target, respectively.
We see from the figures that our method accurately described the target space structure despite its simplicity, using only \(10\%\) of target task training data.
The discovered transferability relationships also align well with common sense.
For example, \textit{trdo} is more transferable to, e.g., \textit{detr}, since they both contain the \textit{truck} class.
\textit{trdo} also transfers well with \textit{auca} because features useful for distinguishing trucks from dogs are likely also helpful for classifying automobiles from cats.

\section{Conclusions}

In this paper, we proposed a simple, alternative view on NNs that turns layers into linear models in feature spaces and showed that they are KMs.
Based on this construction, we presented a modular learning framework for classification that does not require between-module propagation.
Focusing on the two-module instantiation, we proved its optimality, demonstrated that it matches state-of-the-art performance from end-to-end backpropagation on MNIST and CIFAR-10, and showed that it learns powerful classifiers using almost only implicit, more efficient pairwise labels.
This modular learning framework enables fully modularized DL workflows.
We then demonstrated the benefit of such a workflow in a transfer learning setting, where the transferability among 15 binary classification tasks from CIFAR-10 were to be estimated.
Our simple approach accurately described the task space structure using a fraction of target task training data with practically no computation overhead.

Some future work include a rigorous study of the transferability estimation problem and our proposed solution.
Both theoretical and empirical results can be obtained to enhance the understanding of our method.
A more in-depth study on how our modular learning method can help produce better un/semi-supervised learning paradigms is also of interest.
A study on how the proposed proxy objectives behave in practice and if and how network design choices affect our modular learning can be beneficial especially for practitioners.



\section{Acknowledgements}

This work was supported by DARPA (FA9453-18-1-0039) and ONR (N00014-18-1-2306).

\bibliographystyle{IEEEtran}
\bibliography{main}

\clearpage
\appendix

\noindent\textbf{1. Proving Theorem~\ref{th1}}

    \begin{lemma}
        \label{lem1}
        Given an inner product space \(H\), a unit vector \(\mathbf{e}\in H\), and four other vectors \(\mathbf{v}_+, \mathbf{v}_-, \mathbf{v}_+^\star\), and \(\mathbf{v}_-^\star\) with \(\|\mathbf{v}_+\|_H = \|\mathbf{v}_-\|_H = \|\mathbf{v}_+^\star\|_H = \|\mathbf{v}_-^\star\|_H > 0\), where the norm is the canonical norm induced by the inner product, i.e., \(\|\mathbf{u}\|_H^2 := \langle \mathbf{u}, \mathbf{u} \rangle_H, \forall\mathbf{u}\in H\).
        Assume
        \begin{align}
            \|\mathbf{v}_+ - \mathbf{v}_-\|_H \leq \|\mathbf{v}_+^\star - \mathbf{v}_-^\star\|_H.
        \end{align}
        Then there exists a unit vector \(\mathbf{e}^\star\in H\) such that
        \begin{align}
            \langle \mathbf{e}, \mathbf{v}_+ \rangle_H \leq \langle \mathbf{e}^\star, \mathbf{v}_+^\star \rangle_H;\\
            \langle \mathbf{e}, \mathbf{v}_- \rangle_H \geq \langle \mathbf{e}^\star, \mathbf{v}_-^\star \rangle_H.
        \end{align}
    \end{lemma}
    \begin{proof}
        Throughout, we omit the subscript \(H\) on inner products and norms for brevity, which shall cause no ambiguity since no other inner product or norm will be involved in this proof.

        If \(\mathbf{v}_+^\star = \mathbf{v}_-^\star\), then \(\|\mathbf{v}_+^\star - \mathbf{v}_-^\star\| = 0\), which would imply
        \begin{equation}
            \|\mathbf{v}_+ - \mathbf{v}_-\| = 0 \Longleftrightarrow \mathbf{v}_+ = \mathbf{v}_-.
        \end{equation}
        Then we may choose \(\mathbf{e}^\star\) such that \(\langle \mathbf{e}^\star, \mathbf{v}_+^\star \rangle = \langle \mathbf{e}, \mathbf{v}_+ \rangle\) and the result holds trivially. 

        On the other hand, if \(\mathbf{v}_+^\star = -\mathbf{v}_+^\star\), \(\mathbf{e}^\star = \mathbf{v}_+^\star / \|\mathbf{v}_+^\star\|\) would be a valid choice of \(\mathbf{e}^\star\).
        Indeed, by Cauchy-Schwarz,
        \begin{align}
            &\langle \mathbf{e}^\star, \mathbf{v}_+^\star \rangle = \|\mathbf{v}_+^\star\| \geq \langle \mathbf{e}, \mathbf{v}_+ \rangle;\\
            &\langle \mathbf{e}^\star, \mathbf{v}_-^\star \rangle = -\|\mathbf{v}_-^\star\| \leq \langle \mathbf{e}, \mathbf{v}_- \rangle.
        \end{align}
        Therefore, we may assume that \(\mathbf{v}_+^\star \neq \pm\mathbf{v}_-^\star\).

        For two vectors \(\mathbf{a}, \mathbf{b}\), we define the ``angle'' between them, denoted \(\theta \in [0, \pi]\), via \(\cos\theta := \langle\mathbf{a}, \mathbf{b}\rangle / (\|\mathbf{a}\|\|\mathbf{b}\|)\).
        This angle is well-defined since \(\cos\) is injective on \([0, \pi]\).

        Since
        \begin{equation}
            \langle \mathbf{a}, \mathbf{b} \rangle = -\frac{1}{2}\|\mathbf{a} - \mathbf{b}\|^2 + \frac{1}{2}\|\mathbf{a}\|^2 + \frac{1}{2}\|\mathbf{b}\|^2, \forall\mathbf{a}, \mathbf{b},
        \end{equation}
        \(\|\mathbf{v}_+\| = \|\mathbf{v}_-\| = \|\mathbf{v}_+^\star\| = \|\mathbf{v}_-^\star\| > 0\) and \(\|\mathbf{v}_+ - \mathbf{v}_-\| \leq \|\mathbf{v}_+^\star - \mathbf{v}_-^\star\|\) together implies \(\langle \mathbf{v}_+, \mathbf{v}_- \rangle / (\|\mathbf{v}_+\|\|\mathbf{v}_-\|)\geq \langle \mathbf{v}_+^\star, \mathbf{v}_-^\star \rangle / (\|\mathbf{v}_+^\star\|\|\mathbf{v}_-^\star\|)\), which then implies \(\theta\leq\theta^\star\) since \(\cos\) is strictly decreasing on \([0, \pi]\), where \(\theta\) is the angle between \(\mathbf{v}_+, \mathbf{v}_-\) and \(\theta^\star\) is the angle between \(\mathbf{v}_+^\star, \mathbf{v}_-^\star\).

        Let \(\gamma_+\) be the angle between \(\mathbf{e}, \mathbf{v}_+\) and \(\gamma_-\) that between \(\mathbf{e}, \mathbf{v}_-\).
        Note that \(\gamma_- - \gamma_+ \in [0, \pi]\).
        Define
        \begin{align}
            p := \langle \mathbf{e}, \mathbf{v}_+ \rangle = \|\mathbf{v}_+\|\cos\gamma_+;\\
            n := \langle \mathbf{e}, \mathbf{v}_- \rangle = \|\mathbf{v}_-\|\cos\gamma_-.
        \end{align}

        Now, suppose that we have shown
        \begin{align}
            \label{intermediate1}
            &\gamma_- - \gamma_+ \leq\theta;\\
            \label{intermediate2}
            &\exists \mathbf{e}^\star \text{ s.t. } \gamma_-^\star - \gamma_+^\star = \theta^\star \text{ and } \gamma_-^\star = \gamma_-,
        \end{align}
        where \(\gamma_+^\star\) is the angle between \(\mathbf{e}^\star, \mathbf{v}_+^\star\) and \(\gamma_-^\star\) that between \(\mathbf{e}^\star, \mathbf{v}_-^\star\).
        Define:
        \begin{align}
            p^\star := \langle \mathbf{e}^\star, \mathbf{v}_+^\star \rangle = \|\mathbf{v}_+^\star\|\cos\gamma_+^\star;\\
            n^\star := \langle \mathbf{e}^\star, \mathbf{v}_-^\star \rangle = \|\mathbf{v}_-^\star\|\cos\gamma_-^\star.
        \end{align}
        Then using \(\|\mathbf{v}_-\| = \|\mathbf{v}_-^\star\| > 0\) and the earlier result that \(\theta\leq\theta^\star\), we would have
        \begin{align}
            n = n^\star;\\
            \gamma_+ \geq\gamma_+^\star,
        \end{align}
        which, together with the fact that \(\cos\) is strictly decreasing on \([0, \pi]\) and the assumption that \(\|\mathbf{v}_+\| = \|\mathbf{v}_+^\star\| > 0\), implies
        \begin{align}
            n = n^\star;\\
            p \leq p^\star,
        \end{align}
        proving the result.

        To prove Eq.~\ref{intermediate1}, it suffices to show \(\cos(\gamma_- - \gamma_+)\geq\cos\theta\) since \(\cos\) is decreasing on \([0, \pi]\), to which both \(\gamma_- - \gamma_+\) and \(\theta\) belong.
        To this end, we have
        \begin{align}
            \cos(\gamma_- - \gamma_+) &= \cos\gamma_-\cos\gamma_+ + \sin\gamma_-\sin\gamma_+\\
            &= \frac{pn}{\|\mathbf{v}_+\|\|\mathbf{v}_-\|} + \sqrt{\frac{(\|\mathbf{v}_+\|^2 - p^2)(\|\mathbf{v}_-\|^2 - n^2)}{\|\mathbf{v}_+\|^2\|\mathbf{v}_-\|^2}}.
        \end{align}
        Since \(\|\mathbf{v}_+\|^2 - p^2 = \|\mathbf{v}_+\|^2 + p^2 - 2 p^2 = \|\mathbf{v}_+\|^2 + p^2 - 2p \langle \mathbf{e}, \mathbf{v}_+ \rangle = \|\mathbf{v}_+ - p\mathbf{e}\|^2\) and similarly \(\|\mathbf{v}_-\|^2 - n^2 = \|\mathbf{v}_- - n\mathbf{e}\|^2\),
        \begin{align}
            \cos(\gamma_- - \gamma_+) &= \frac{pn + \|\mathbf{v}_+ - p\mathbf{e}\|\|\mathbf{v}_- - n\mathbf{e}\|}{\|\mathbf{v}_+\|\|\mathbf{v}_-\|}\\
            &\geq \frac{pn + \langle \mathbf{v}_+ - p\mathbf{e}, \mathbf{v}_- - n\mathbf{e} \rangle}{\|\mathbf{v}_+\|\|\mathbf{v}_-\|}\\
            &= \frac{\langle \mathbf{v}_+, \mathbf{v}_- \rangle}{\|\mathbf{v}_+\|\|\mathbf{v}_-\|}\\
            &=\cos\theta,
        \end{align}
        where the inequality is due to Cauchy-Schwarz.

        To prove~\ref{intermediate2}, it suffices to show that there exists \(\mathbf{e}^\star\) such that
        \begin{enumerate}[label=\Alph*]
            \item one of \(\mathbf{v}_+^\star - p^\star\mathbf{e}^\star, \mathbf{v}_-^\star - n^\star\mathbf{e}^\star\) is a scalar multiple of the other;\label{i2equiv1}
            \item \(\cos\gamma_-^\star = \cos\gamma_-\).\label{i2equiv2}
        \end{enumerate}

        Indeed, \ref{i2equiv1} implies \(\|\mathbf{v}_+^\star - p^\star\mathbf{e}^\star\|\|\mathbf{v}_-^\star - n^\star\mathbf{e}^\star\| = \langle \mathbf{v}_+^\star - p^\star\mathbf{e}^\star, \mathbf{v}_-^\star - n^\star\mathbf{e}^\star \rangle\), which, together with similar arguments as we used to prove Eq.~\ref{intermediate1}, implies \(\gamma_-^\star - \gamma_+^\star = \theta^\star\).
        Also note that \ref{i2equiv2} is equivalent to \(n = n^\star\).

        We prove existence constructively.
        Specifically, we set \(\mathbf{e}^\star = a\mathbf{v}_+^\star + b\mathbf{v}_-^\star, a, b\in\mathbb{R}\) and find \(a, b\) such that \(\mathbf{e}^\star\) satisfies \ref{i2equiv1} and \ref{i2equiv2} simultaneously.

        Let \(s = \langle \mathbf{v}_+^\star, \mathbf{v}_-^\star \rangle, r = \|\mathbf{v}_+^\star\|^2 = \|\mathbf{v}_-^\star\|^2\).
        Note that we immediately have:
        \begin{equation}
            r > 0; |s| \leq r.
        \end{equation}
        The assumption \(\mathbf{v}_+^\star\neq\pm\mathbf{v}_-^\star\) implies \(|s| < r\).

        Since \(\mathbf{e}^\star\) is a unit vector, we have the following constraint on \(a, b\):
        \begin{equation}
            \label{s}
            a^2 r + b^2 r + 2abs = 1.
        \end{equation}

        And some simple algebra yields
        \begin{align}
            p^\star = ar + bs;\\
            n^\star = as + br.
        \end{align}

        To identify those \(a, b\) such that \ref{i2equiv1} holds, we first rewrite \(\mathbf{v}_+^\star - p^\star\mathbf{e}^\star\) and \(\mathbf{v}_-^\star - n^\star\mathbf{e}^\star\) as follows.
        \begin{align}
            \mathbf{v}_+^\star - p^\star\mathbf{e}^\star &= \mathbf{v}_+^\star - (ar + bs) (a\mathbf{v}_+^\star + b\mathbf{v}_-^\star)\\
            &= (1 - a^2r -abs)\mathbf{v}_+^\star + (-abr - b^2s)\mathbf{v}_-^\star.
        \end{align}
        Similarly,
        \begin{equation}
            \mathbf{v}_-^\star - n^\star\mathbf{e}^\star = (- a^2s -abr)\mathbf{v}_+^\star + (1 -abs - b^2r)\mathbf{v}_-^\star.
        \end{equation}

        Define
        \begin{align}
           &w_{1, +} := 1 - a^2 r - abs;\\
           &w_{1, -} := -abr - b^2 s;\\
           &w_{2, +} := -a^2 s - abr;\\
           &w_{2, -} := 1 - abs - b^2 r.
        \end{align}
        Then we have
        \begin{align}
            &\mathbf{v}_+^\star - p^\star\mathbf{e}^\star = w_{1, +}\mathbf{v}_+^\star + w_{1, -}\mathbf{v}_-^\star;\\
            &\mathbf{v}_-^\star - n^\star\mathbf{e}^\star = w_{2, +}\mathbf{v}_+^\star + w_{2, -}\mathbf{v}_-^\star.
        \end{align}

        Assuming none of \(w_{1, +}, w_{1, -}, w_{2, +}, w_{2, -}\) is \(0\), \ref{i2equiv1} is equivalent to
        \begin{equation}
            w_{1, +}w_{2, -} = w_{1, -}w_{2, +}.
        \end{equation}
        To check that this is always true, we have
        \begin{align}
            &w_{1, +}w_{2, -} \\
            &\quad= a^3 brs + ab^3 rs + a^2 b^2 s^2 + a^2 b^2 r^2 - 2abs - r(a^2 + b^2) + 1\\
            &\quad =a^3 brs + ab^3 rs + a^2 b^2 s^2 + a^2 b^2 r^2
        \end{align}
        because of Eq.~\ref{s}.
        And
        \begin{align}
            w_{1, -}w_{2, +} &= a^3 brs + ab^3 rs + a^2 b^2 s^2 + a^2 b^2 r^2.
        \end{align}
        Therefore, \ref{i2equiv1} is always true.

        If at least one of \(w_{1, +}, w_{1, -}, w_{2, +}, w_{2, -}\) is \(0\), we have the following mutually exclusive cases:
        \begin{enumerate}[label=\roman*]
            \item\label{case0} one of the four coefficients is \(0\) while the other three are not; \quad\xmark
            \item \(w_{1, +} = w_{1, -} = 0\), the others are not \(0\); \quad\cmark
            \item \(w_{2, +} = w_{2, -} = 0\), the others are not \(0\); \quad\cmark
            \item\label{case1} \(w_{1, +} = w_{2, -} = 0\), the others are not \(0\); \quad\xmark
            \item\label{case2} \(w_{1, -} = w_{2, +} = 0\), the others are not \(0\); \quad\xmark
            \item  \(w_{1, +} = w_{2, +} = 0\), the others are not \(0\); \quad\cmark
            \item \(w_{1, -} = w_{2, -} = 0\), the others are not \(0\); \quad\cmark
            \item three of the four coefficients are \(0\) while one is not; \quad\cmark
            \item \(w_{1, +} = w_{1, -} = w_{2, +} = w_{2, -} = 0\), \quad\cmark
        \end{enumerate}
        where the cases marked with \(\xmark\) are the ones where \ref{i2equiv1} cannot be true (so our choice of \(a, b\) cannot fall into these cases) and the ones marked with \(\cmark\) are the ones where \ref{i2equiv1} is true.
        Note that \ref{i2equiv1} cannot be true in cases~\ref{case1} and~\ref{case2} because if \ref{i2equiv1} was true, it would imply that \(\mathbf{v}_+^\star\) and \(\mathbf{v}_-^\star\) are linearly dependent.
        However, since \(\|\mathbf{v}_+^\star\| = \|\mathbf{v}_-^\star\|\), we would have either \(\mathbf{v}_+^\star = \mathbf{v}_-^\star\) or \(\mathbf{v}_+^\star = -\mathbf{v}_-^\star\), both of which have been excluded from the discussion in the beginning of this proof.

        Therefore, \ref{i2equiv1} is satisfied by any \(a, b\) that satisfy Eq.~\ref{s} but none of case~\ref{case0}, \ref{case1}, and \ref{case2}.

        We now turn to the search for \(a, b\) such that \ref{i2equiv2} holds.
        \ref{i2equiv2} is equivalent to
        \begin{equation}
            as + br = n \Longleftrightarrow b = \frac{1}{r}(n - as).
        \end{equation}

        Therefore, finding \(a, b\) such that \ref{i2equiv1} and \ref{i2equiv2} hold simultaneously amounts to finding \(a, b\) such that
        \begin{align}
            \label{final1}
            &b =\frac{1}{r}(n - as);\\
            \label{final2}
            &a^2 r + b^2 r + 2abs = 1;\\
            \label{final3}
            &\text{none of case~\ref{case0}, \ref{case1}, \ref{case2} is true.}
        \end{align}

        Now, substituting \(b = (n - as) / r\) into Eq.~\ref{final2} and solving for \(a\), we have
        \begin{equation}
            a^2 = \frac{r - n^2}{r^2 - s^2},
        \end{equation}
        and we choose
        \begin{equation}
            a = \sqrt{\frac{r - n^2}{r^2 - s^2}}.
        \end{equation}
        This root is real since \(n = \langle \mathbf{e}, \mathbf{v}_- \rangle\) and therefore \(n^2 = r\cos^2\gamma_- \leq r\).

        To verify that
        \begin{equation}
            a = \sqrt{\frac{r - n^2}{r^2 - s^2}}, b = \frac{1}{r}(n - as).
            \label{rewrite}
        \end{equation}
        satisfies~\ref{final3}, first note that since this solution of \(a, b\) comes from solving \(n = n^\star\) and Eq.~\ref{final2}, we have
        \begin{align}
            &w_{1, +} = 1 - a(bs + ar) = b(br + as) = bn;\\
            &w_{1, -} \propto b(bs + ar);\\
            &w_{2, +} \propto a(as + br) = an;\\
            &w_{2, -} = 1 - b(as + br) = 1 - bn.
        \end{align}

        We now analyze each one of case~\ref{case0}, \ref{case1}, and \ref{case2} individually and show that our choice of \(a, b\) does not fall into any of them, proving that this particular \(a, b\) satisfies Eq.~\ref{final1}, \ref{final2}, and condition~\ref{final3}.

        \noindent\textbf{case~\ref{case0}:}

        If \(w_{1, +} = 0\), then either \(b = 0\) or \(n = 0\), resulting in \(w_{1, -} \propto b(bs + ar) = 0\) or \(w_{2, +} \propto an = 0\), i.e., there must be at least \(2\) coefficients being \(0\).

        If \(w_{1, -} = 0\), then either \(b = 0\), in which case \(w_{1, +} = 0\), or \(bs + ar = 0\).
        In the latter case, we would then use Eq.~\ref{final2} and have
        \begin{align}
            abs + a^2 r = 0 \Longrightarrow b^2 r + abs = 1 \Longrightarrow b(br + as) = bn = 1 \nonumber\\
            \Longrightarrow w_{2, -} = 1 - bn = 0.
        \end{align}
        Again, we would have at least \(2\) nonzero coefficients.

        If \(w_{2, +} = 0\), then either \(n = 0\), which would result in \(w_{1, +} = bn = 0\), or \(a = 0\).
        Assuming \(a = 0\), Eq.~\ref{rewrite} would imply \(n = \pm\sqrt{r}\) and \(b = n / r\).
        Either way, we would have \(b = 1 / n\) and therefore \(w_{2, -} = 1 - bn = 0\).

        Finally, if \(w_{2, -} = 0\), then first assuming \(n\neq 0\), we would have \(b = 1 / n\).
        Then Eq.~\ref{rewrite} would give \(1/n = (n - as)/r\).
        Solving for \(a\), we have \(a = (n^2 - r)/(ns)\).
        \(r\geq n^2\) would imply that \(a \leq 0\).
        On the other hand, Eq.~\ref{rewrite} also implied \(a \geq 0\).
        Hence, \(a\) must be \(0\), in which case \(w_{2, +} \propto an = 0\).
        If \(n = 0\), we would have \(w_{2, +} = 0\) as well.

        \noindent\textbf{case~\ref{case1}:}

        It is easy to see that this case is impossible since \(w_{1, +} + w_{2, -} = 1\).

        \noindent\textbf{case~\ref{case2}:}

        If \(w_{1, -} = w_{2, +} = 0\), then we have already shown in the proof of case~\ref{case0} that either \(w_{1, +}\) or \(w_{2, -}\) must be \(0\), that is, at least \(3\) coefficients would be \(0\).

        In summary, we have found an \(\mathbf{e}^\star\) such that \ref{i2equiv1} and \ref{i2equiv2} hold simultaneously, proving~\ref{intermediate2}.
    \end{proof}

\bigskip

We now prove Theorem~\ref{th1}.

    \begin{proof}
        The result amounts to proving
        \begin{equation}
            L(f_2^\star\circ F_2^\star, S)\leq L(f_2\circ F_1, S), \forall f_2, F_1.
        \end{equation}

        Define \(S_+\) to be the set of all \(\mathbf{x}_i\) such that \(i \in I_+\) and \(S_-\) the set of all \(\mathbf{x}_j\) such that \(j \in I_-\).
        Let \(\kappa = \frac{1}{n}\sum_{i=1}^n\mathbbm{1}_{\{i \in I_+\}}\), we have
        \begin{align}
            &L(f_2\circ F_1^\star, S) \nonumber\\
            &\quad\leq\kappa\ell_+\left( f_2\circ F_1^\star(\mathbf{x}_+^\star) \right) + (1 - \kappa)\ell_-\left( f_2\circ F_1^\star(\mathbf{x}_-^\star) \right) \nonumber\\
            &\quad\quad+ \lambda g(\|\mathbf{w}\|)\\
            &\quad=\kappa\ell_+\left( \left\langle \frac{\mathbf{w}}{\|\mathbf{w}\|}, \Phi_+^\star \right\rangle\|\mathbf{w}\| + b \right)\\\nonumber
            &\quad\quad+ (1 - \kappa)\ell_-\left( \left\langle \frac{\mathbf{w}}{\|\mathbf{w}\|}, \Phi_-^\star \right\rangle\|\mathbf{w}\| + b \right) + \lambda g(\|\mathbf{w}\|)
        \end{align}
        for some \(\mathbf{x}_+^\star, \mathbf{x}_-^\star\) from \(S_+, S_-\), respectively, where \(\Phi_+^\star := \Phi\left( F_1^\star(\mathbf{x}_+^\star) \right)\) and \(\Phi_-^\star := \Phi\left( F_1^\star(\mathbf{x}_-^\star) \right)\).

    For any \(f_2^\prime\circ F_1^\prime\), let \(f_2^\prime\) be parameterized by \(\mathbf{w}^\prime, b^\prime\).
    We have
    \begin{align}
        &L(f_2^\prime\circ F_1^\prime, S)\nonumber\\
        &\quad\geq\kappa\ell_+\left( f_2^\prime\circ F_1^\prime(\mathbf{x}_+^\prime) \right) + (1 - \kappa)\ell_-\left( f_2^\prime\circ F_1^\prime(\mathbf{x}_-^\prime) \right) \nonumber\\
        &\quad\quad+ \lambda g(\|\mathbf{w}^\prime\|)\\
        &\quad=\kappa\ell_+\left( \left\langle \frac{\mathbf{w}^\prime}{\|\mathbf{w}^\prime\|}, \Phi_+^\prime \right\rangle\|\mathbf{w}^\prime\| + b^\prime \right)\\\nonumber
        &\quad\quad+ (1 - \kappa)\ell_-\left( \left\langle \frac{\mathbf{w}^\prime}{\|\mathbf{w}^\prime\|}, \Phi_-^\prime \right\rangle\|\mathbf{w}^\prime\| + b^\prime \right) + \lambda g(\|\mathbf{w}^\prime\|)
    \end{align}
    for \(\mathbf{x}_+^\prime, \mathbf{x}_-^\prime\) with \(\mathbf{x}_+^\prime\) maximizing \(f_2^\prime\circ F_1^\prime(\mathbf{x}_i)\) over \(\mathbf{x}_i\in S_+\) and \(\mathbf{x}_-^\prime\) minimizing \(f_2^\prime\circ F_1^\prime(\mathbf{x}_j)\) over \(\mathbf{x}_j\in S_-\), where \(\Phi_+^\prime := \Phi\left( F_1^\prime(\mathbf{x}_+^\prime) \right)\) and \(\Phi_-^\prime := \Phi\left( F_1^\prime(\mathbf{x}_-^\prime) \right)\).

    Using the assumption on \(F_1^\star\),
        \begin{equation}
            \|\Phi_+^\star - \Phi_-^\star\| \geq \|\Phi_+^\prime - \Phi_-^\prime\|.
        \end{equation}

    Then using the earlier Lemma, there exists a unit vector \(\mathbf{e}^\star\) such that
       \begin{align}
            \left\langle \mathbf{e}^\star, \Phi_+^\star \right\rangle \geq \left\langle \frac{\mathbf{w}^\prime}{\|\mathbf{w}^\prime\|}, \Phi_+^\prime \right\rangle;\\
            \left\langle \mathbf{e}^\star, \Phi_-^\star \right\rangle \leq \left\langle \frac{\mathbf{w}^\prime}{\|\mathbf{w}^\prime\|}, \Phi_-^\prime \right\rangle.\\
       \end{align}
    Let \(A := \{\mathbf{w}: \|\mathbf{w}\| = \|\mathbf{w}^\prime\|\}\), then evidently, \(\mathbf{e}^\star\|\mathbf{w}^\prime\|\in A\), and we have,
    \begin{align}
        &L(f_2^\prime\circ F_1^\prime, S)\nonumber\\
        &\quad\geq\kappa\ell_+\left( \left\langle \mathbf{e}^\star, \Phi_+^\star \right\rangle\|\mathbf{w}^\prime\| + b^\prime \right)\\\nonumber
        &\quad\quad+ (1 - \kappa)\ell_-\left( \left\langle \mathbf{e}^\star, \Phi_-^\star \right\rangle\|\mathbf{w}^\prime\| + b^\prime \right) + \lambda g\left(\|\mathbf{w}^\prime\|\right)\\
        &\quad=\kappa\ell_+\left( \left\langle \frac{\mathbf{e}^\star\|\mathbf{w}^\prime\|}{\left\|\mathbf{e}^\star\|\mathbf{w}^\prime\|\right\|}, \Phi_+^\star \right\rangle\left\|\mathbf{e}^\star\|\mathbf{w}^\prime\|\right\| + b^\prime \right)\\\nonumber
        &\quad\quad+ (1 - \kappa)\ell_-\left( \left\langle \frac{\mathbf{e}^\star\|\mathbf{w}^\prime\|}{\left\|\mathbf{e}^\star\|\mathbf{w}^\prime\|\right\|}, \Phi_-^\star \right\rangle\left\|\mathbf{e}^\star\|\mathbf{w}^\prime\|\right\| + b^\prime \right) \nonumber\\
        &\quad\quad+ \lambda g\left(\left\|\mathbf{e}^\star\|\mathbf{w}^\prime\|\right\|\right)\\
        &\quad\geq\min_{\mathbf{w}\in A}\kappa\ell_+\left( \left\langle \frac{\mathbf{w}}{\|\mathbf{w}\|}, \Phi_+^\star \right\rangle\|\mathbf{w}\| + b^\prime \right)\\\nonumber
        &\quad\quad+ (1 - \kappa)\ell_-\left( \left\langle \frac{\mathbf{w}}{\|\mathbf{w}\|}, \Phi_-^\star \right\rangle\|\mathbf{w}\| + b^\prime \right) + \lambda g\left(\|\mathbf{w}\|\right)\\
        &\quad\geq\min_{\mathbf{w}\in A, b}\kappa\ell_+\left( \left\langle \frac{\mathbf{w}}{\|\mathbf{w}\|}, \Phi_+^\star \right\rangle\|\mathbf{w}\| + b \right)\\\nonumber
        &\quad\quad+ (1 - \kappa)\ell_-\left( \left\langle \frac{\mathbf{w}}{\|\mathbf{w}\|}, \Phi_-^\star \right\rangle\|\mathbf{w}\| + b \right) + \lambda g\left(\|\mathbf{w}\|\right)\\
        &\quad\geq\min_{\mathbf{w}\in A, b} L(f_2\circ F_1^\star, S)\\
        &\quad\geq\min_{\mathbf{w}, b}L(f_2\circ F_1^\star, S)\\
        &\quad= L(f_2^\star\circ F_1^\star, S).
    \end{align}
    This proves the result.
    \end{proof}

\bigskip

\noindent\textbf{2. Example Loss Functions Satisfying Conditions of Theorem~\ref{th1}}

The empirical risk of many popular loss functions can be decomposed into two terms with one nondecreasing and the other nonincreasing such that the condition required by Theorem~\ref{th1} is satisfied.

\begin{itemize}
    \item softmax + cross-entropy (two-class version):
    \begin{align}
        &\ell(F, S) = \frac{1}{n}\sum_{i=1}^n -\mathbbm{1}_{\{i\in I_+\}} \ln\left( \sigma\left( F(\mathbf{x}_i) \right) \right) \\\nonumber
        &\quad\quad\quad- \mathbbm{1}_{\{i\in I_-\}}\ln\left( 1 - \sigma\left( F(\mathbf{x}_i) \right) \right) \\
        &= \frac{1}{n}\sum_{i\in I_+}\ln\left(e^{-F(\mathbf{x}_i)} + 1\right) + \frac{1}{n}\sum_{j\in I_-}\ln\left(e^{F(\mathbf{x}_j)} + 1\right),
    \end{align}
    where \(\sigma\) is the softmax nonlinearity.
    \item tanh + mean squared error (this decomposition works for any monotonic nonlinearity with the value of \(y_i\) adjusted for the range of the nonlinearity):
    \begin{align}
        &\ell(F, S) = \frac{1}{n}\sum_{i=1}^n \left(y_i - \delta\left(F(\mathbf{x}_i)\right)\right)^2\\
        &\,=\frac{1}{n}\sum_{i\in I_+}\left(1 - \delta\left(F(\mathbf{x}_i)\right)\right)^2 + \frac{1}{n}\sum_{j\in I_-}\left(1 + \delta\left(F(\mathbf{x}_j)\right)\right)^2
    \end{align}
    where \(\delta\) is the hyperbolic tangent nonlinearity.
    \item hinge loss:
    \begin{align}
        &\ell(F, S) = \frac{1}{n}\sum_{i=1}^n\max(0, 1 - y_iF(\mathbf{x}_i))\\
        &\,= \frac{1}{n}\sum_{i\in I_+}\max(0, 1 - F(\mathbf{x}_i)) + \frac{1}{n}\sum_{j\in I_-}\max(0, 1 + F(\mathbf{x}_j)).
    \end{align}
\end{itemize}

\bigskip

\noindent\textbf{3. Additional Transferability Results}
\begin{figure*}[t]
    \centering
    \subfloat{
        \includegraphics[width=\columnwidth]{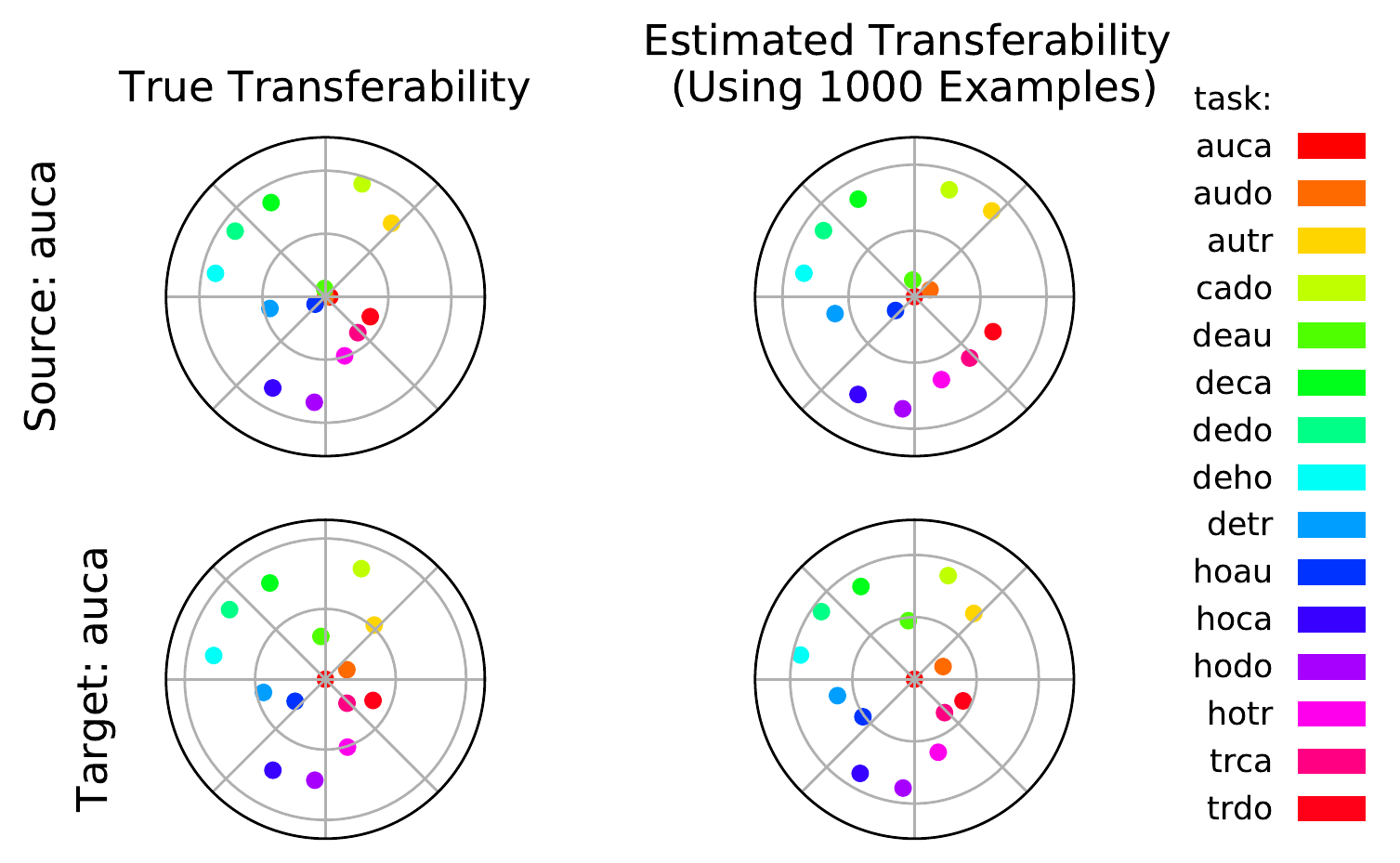}
    }
    \subfloat{
        \includegraphics[width=\columnwidth]{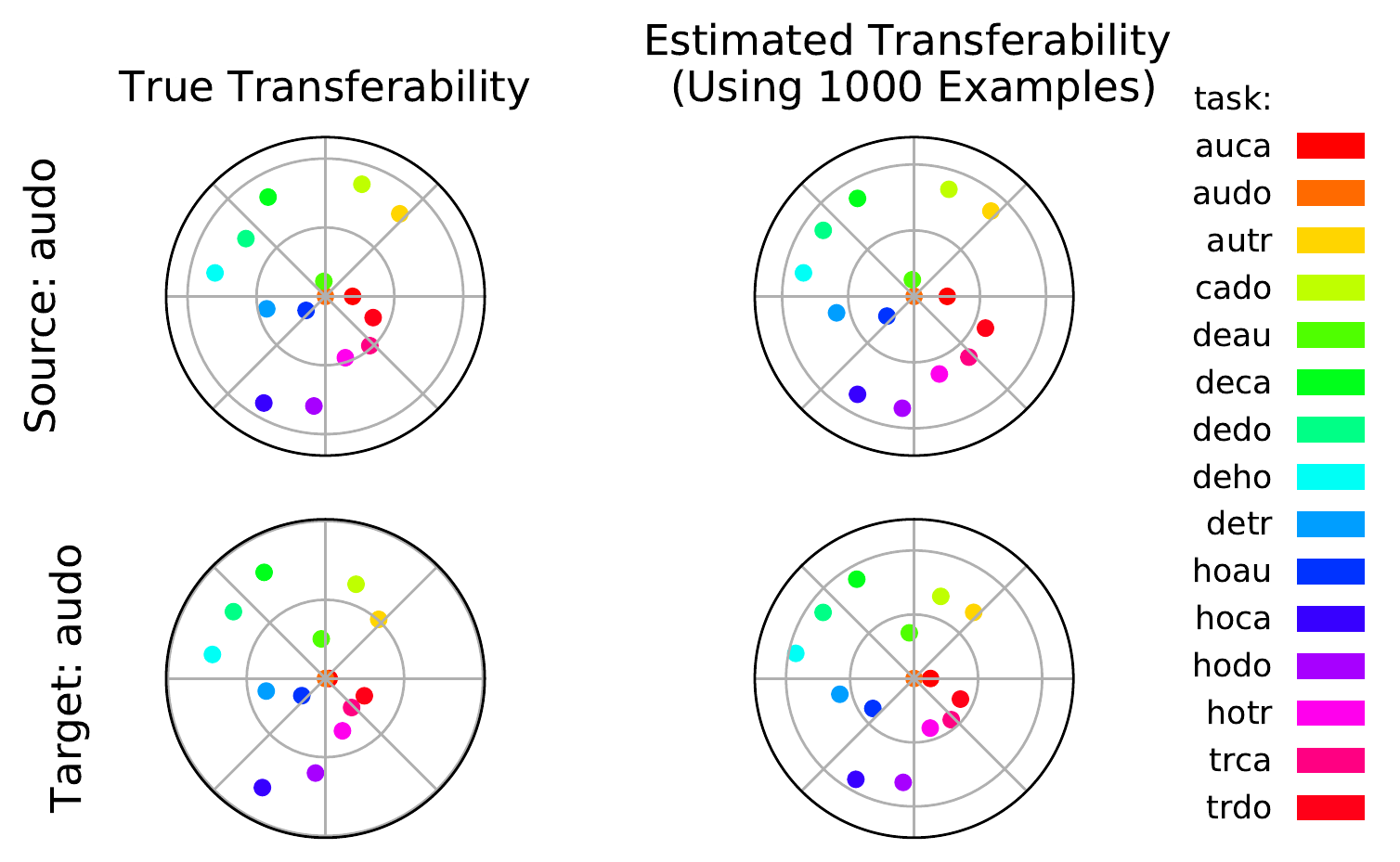}
    }
    \vspace{-10pt}
    \subfloat{
        \includegraphics[width=\columnwidth]{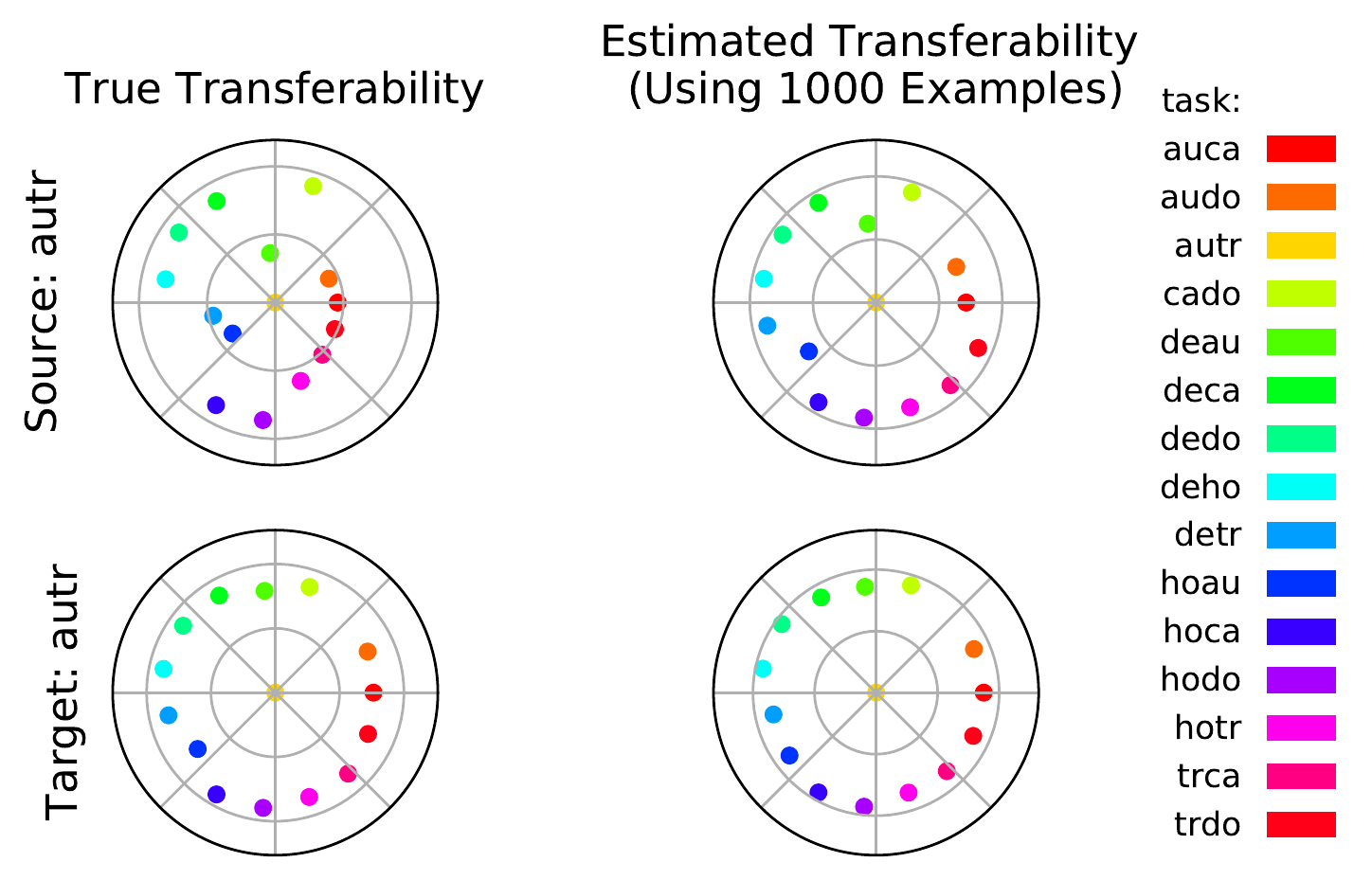}
    }
    \subfloat{
        \includegraphics[width=\columnwidth]{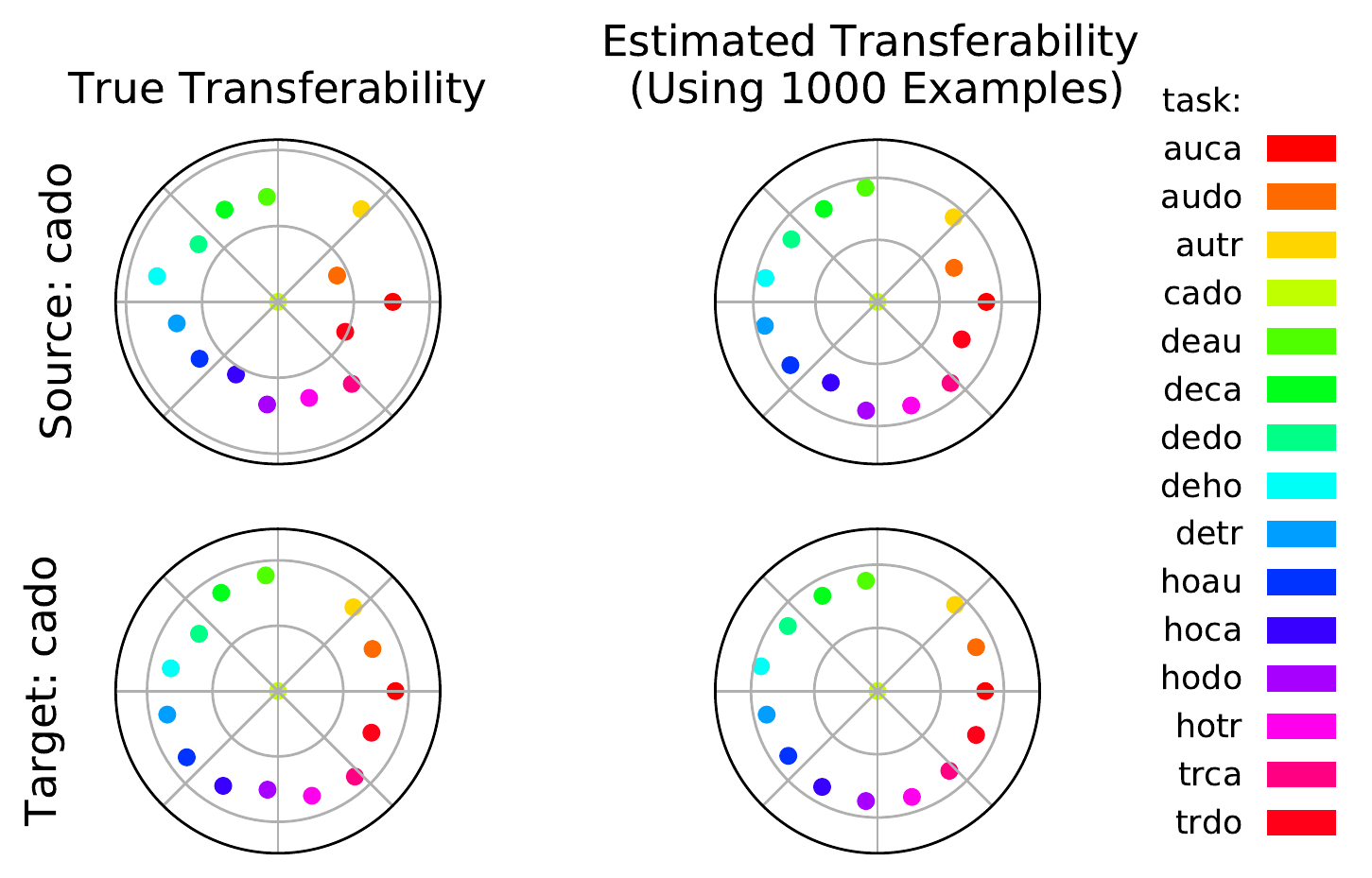}
    }
    \vspace{-10pt}
    \subfloat{
        \includegraphics[width=\columnwidth]{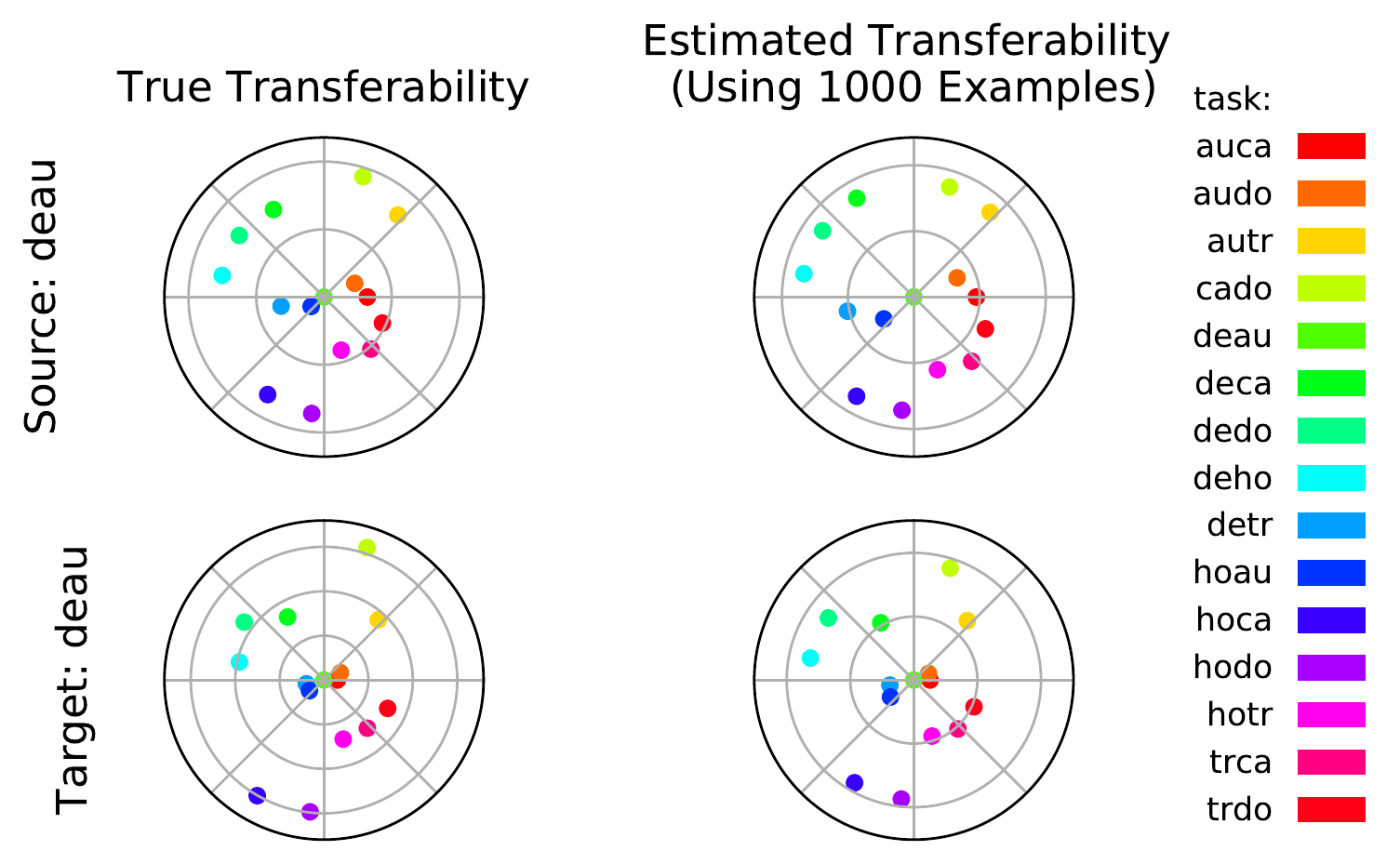}
    }
    \subfloat{
        \includegraphics[width=\columnwidth]{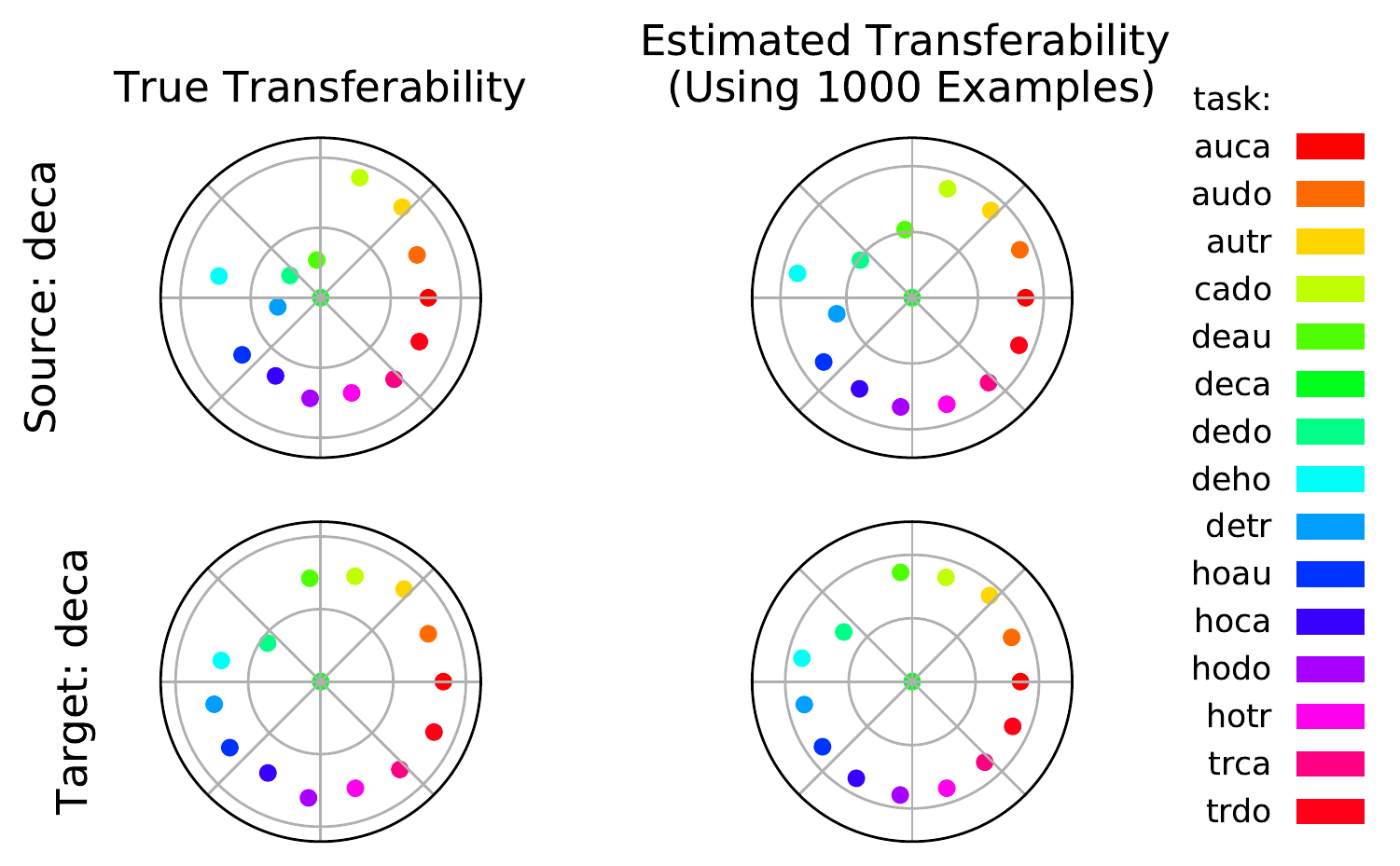}
    }
    \vspace{-10pt}
    \subfloat{
        \includegraphics[width=\columnwidth]{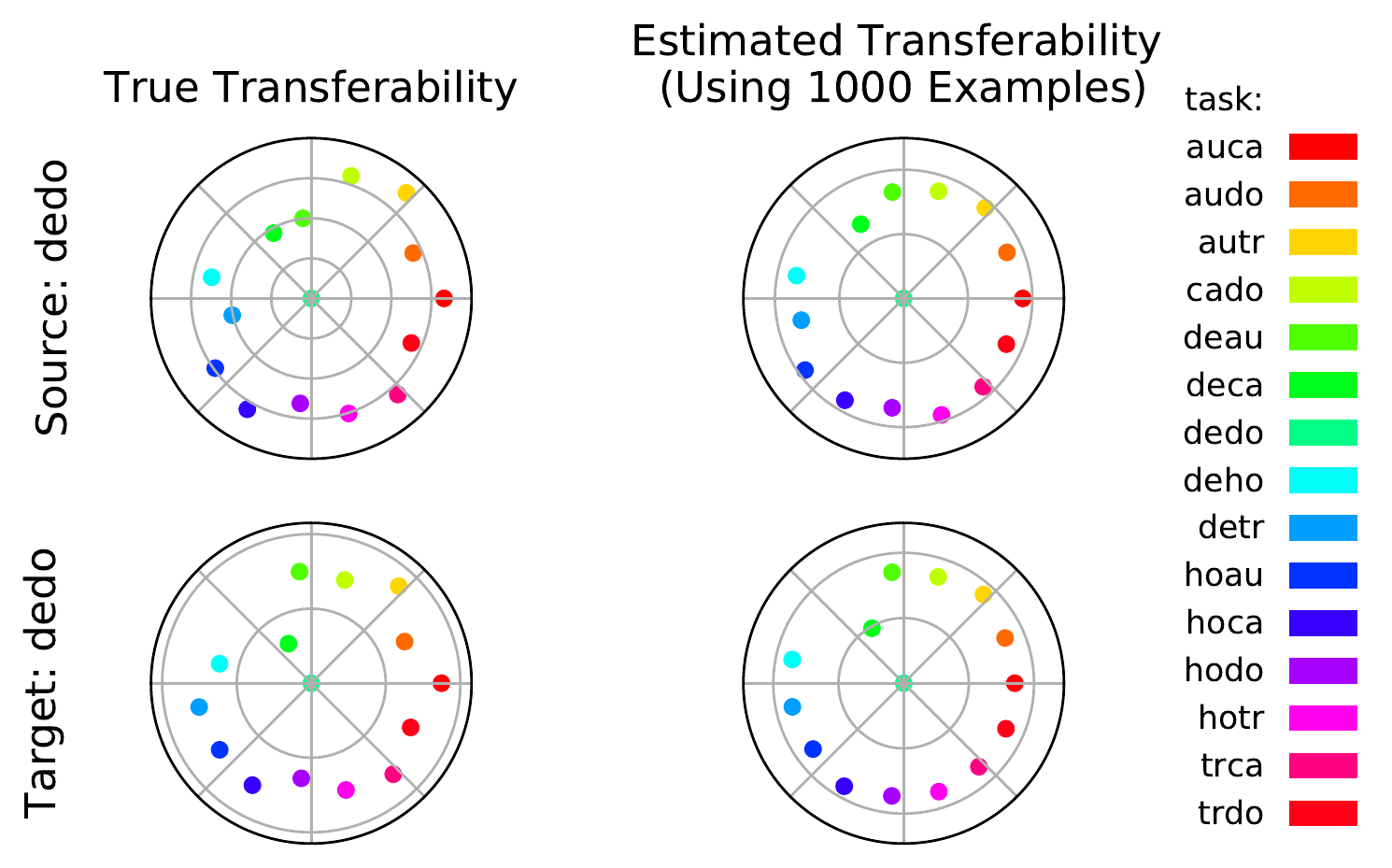}
    }
    \subfloat{
        \includegraphics[width=\columnwidth]{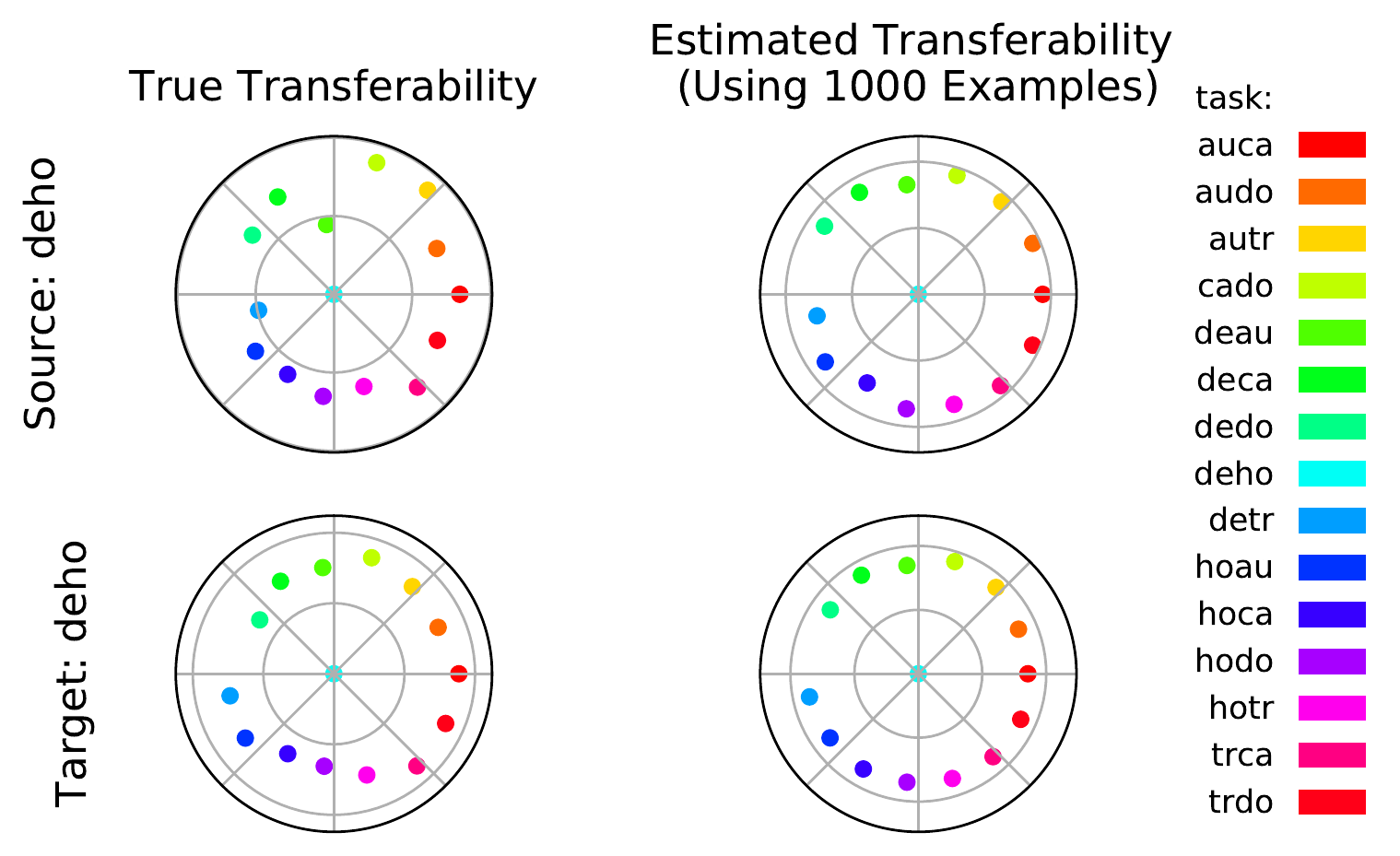}
    }
    \caption{
    Additional figures with other tasks as the source/target task, supplementing Fig.~\ref{fig6}.
    }
    \label{fig7}
\end{figure*}
\begin{figure*}[t]
    \centering
    \subfloat{
        \includegraphics[width=\columnwidth]{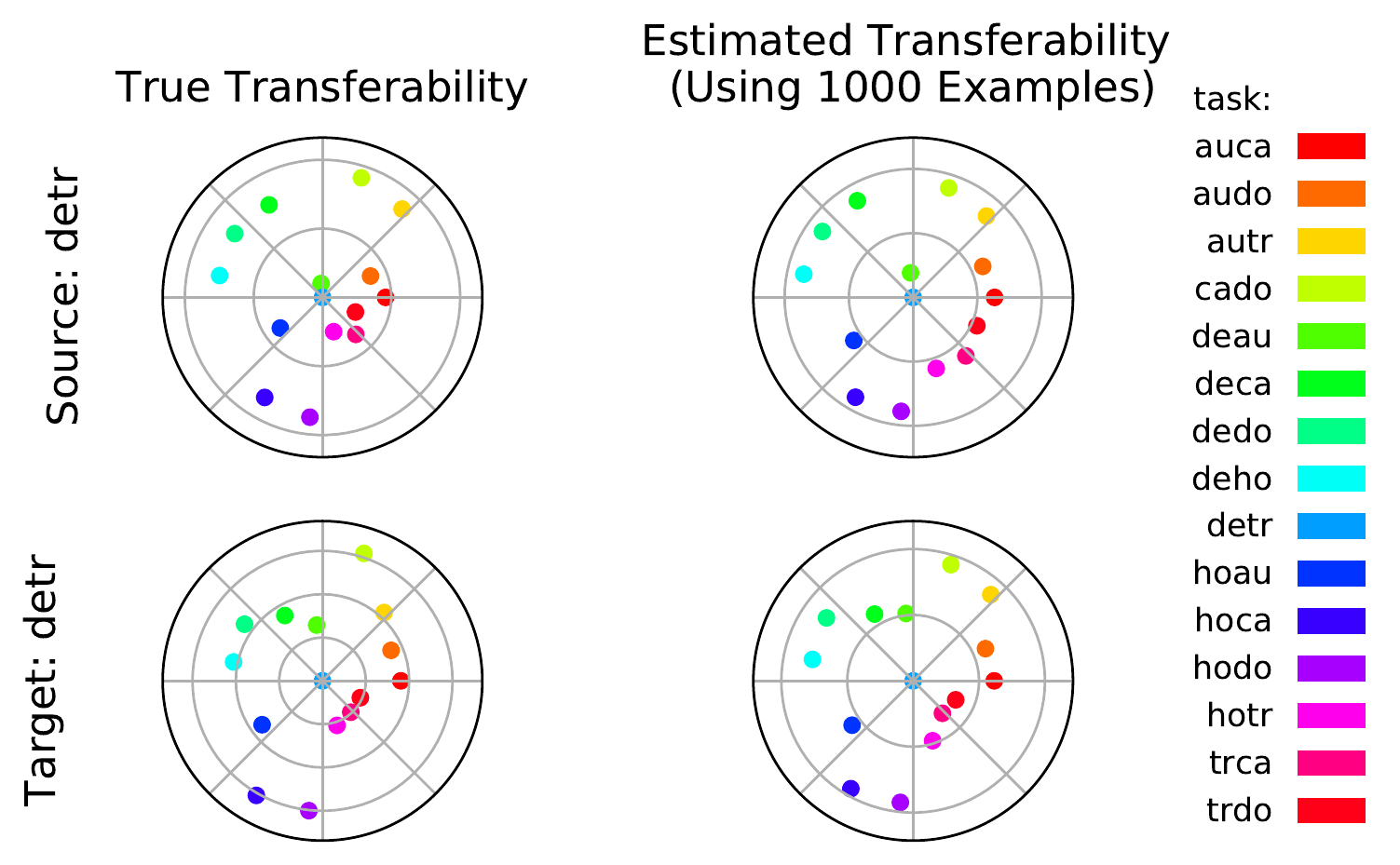}
    }
    \subfloat{
        \includegraphics[width=\columnwidth]{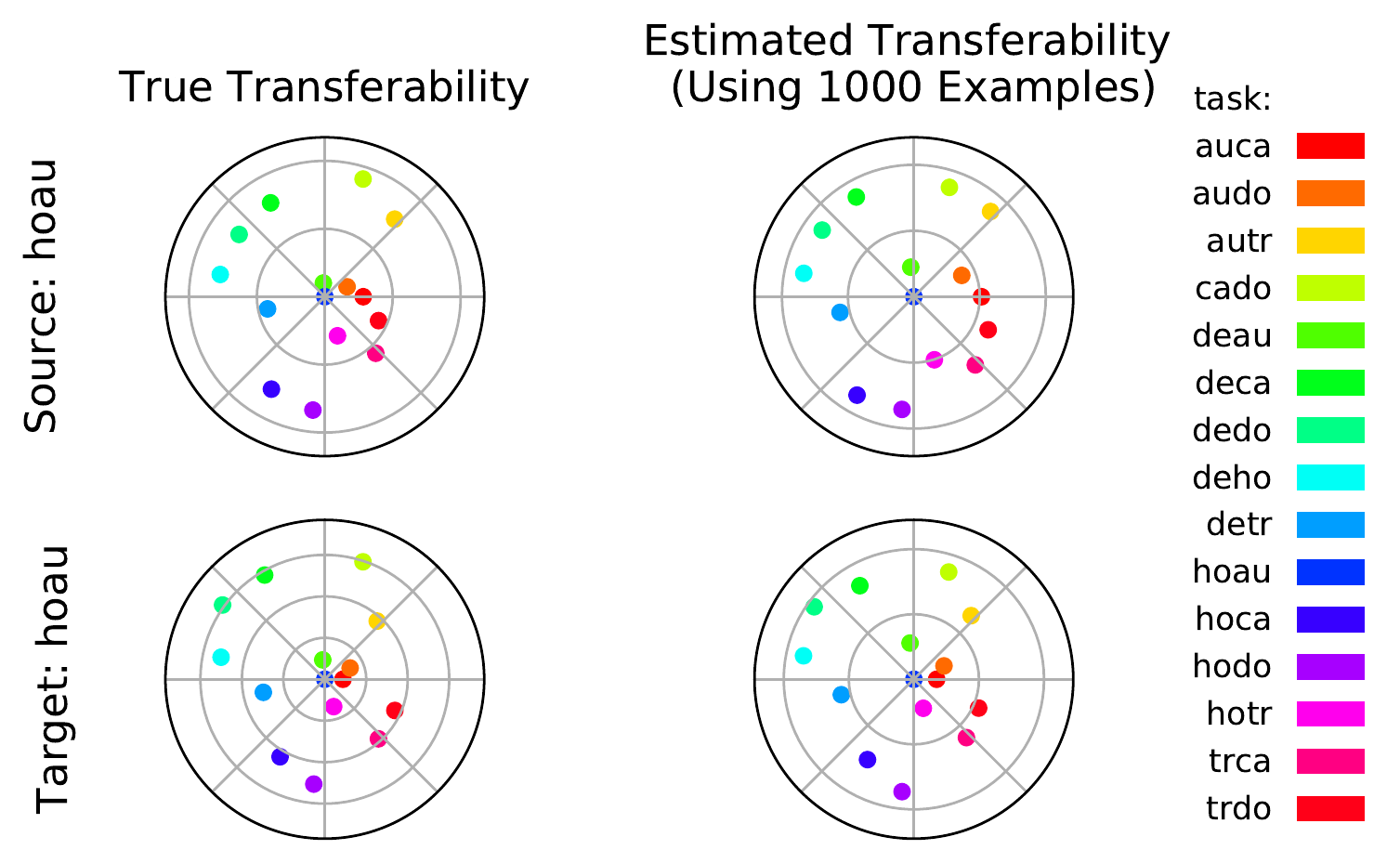}
    }
    \vspace{-10pt}
    \subfloat{
        \includegraphics[width=\columnwidth]{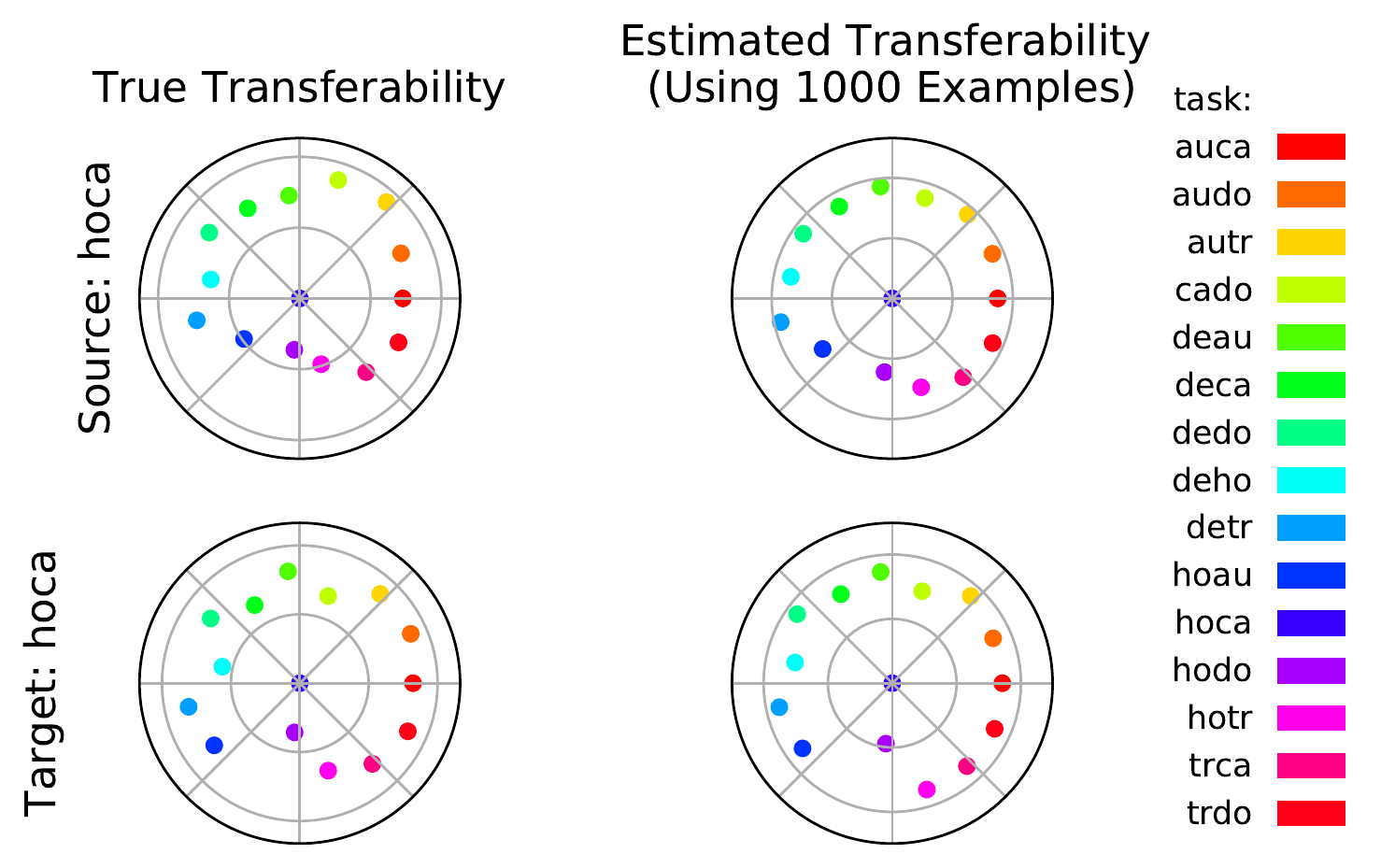}
    }
    \subfloat{
        \includegraphics[width=\columnwidth]{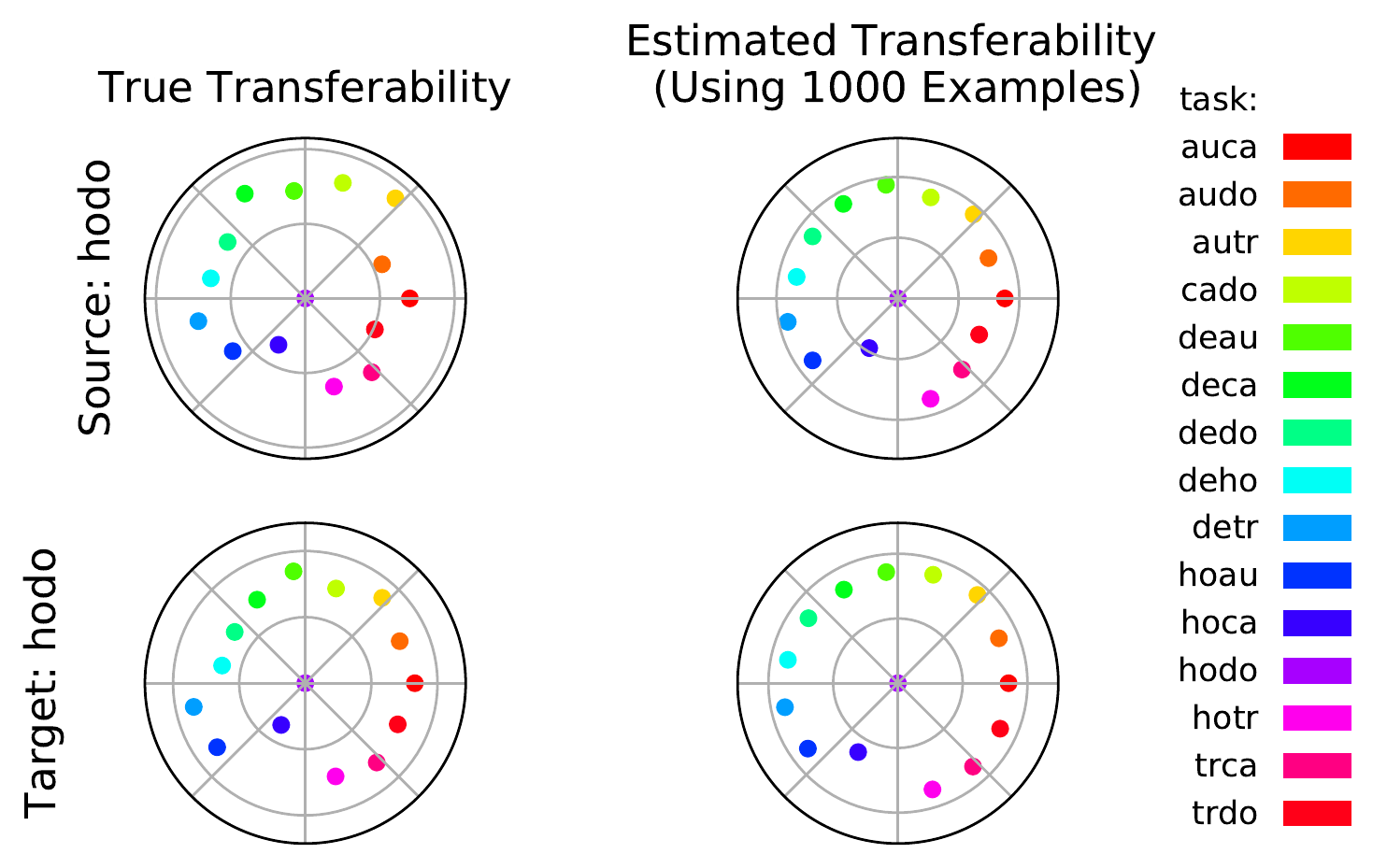}
    }
    \vspace{-10pt}
    \subfloat{
        \includegraphics[width=\columnwidth]{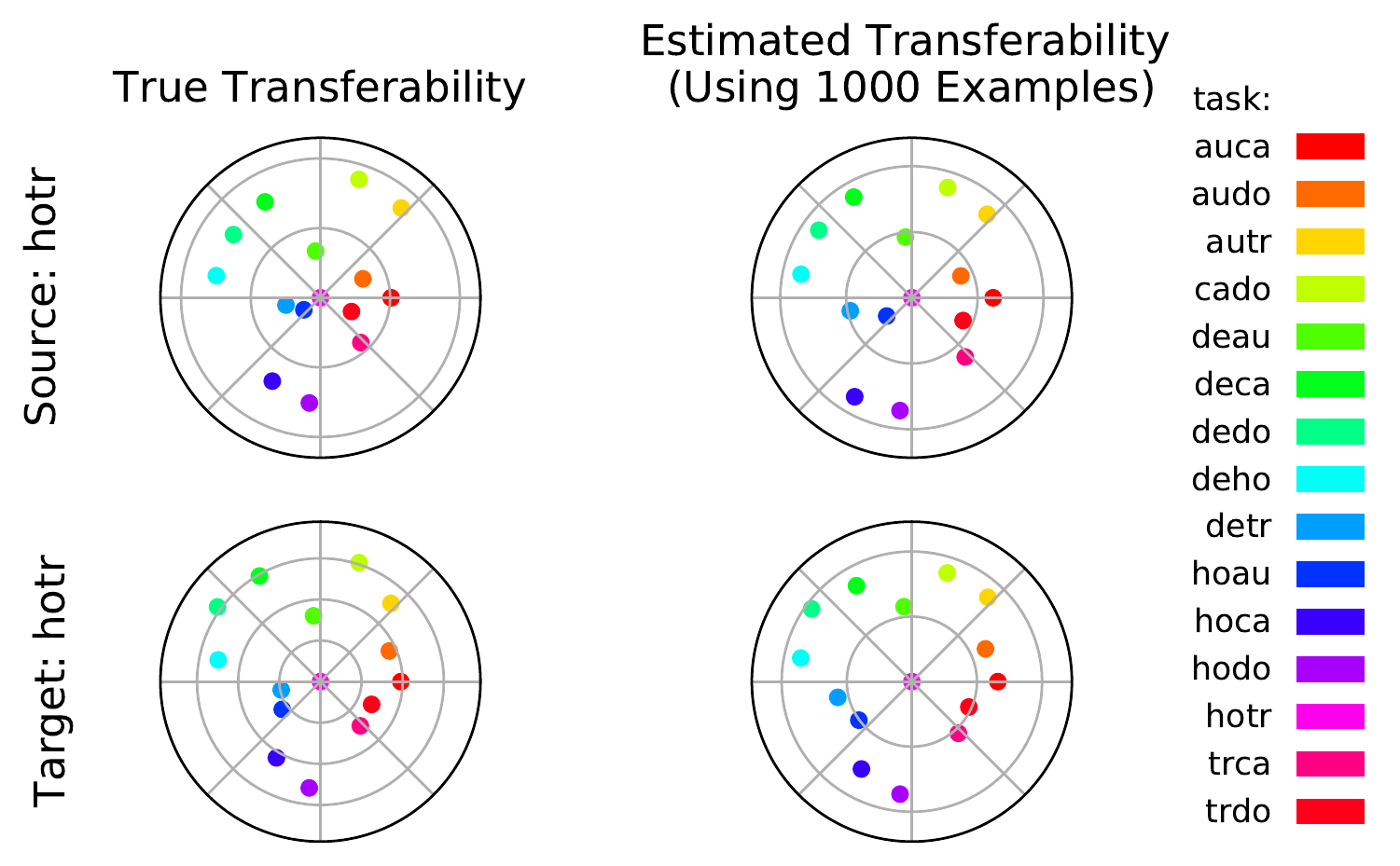}
    }
    \subfloat{
        \includegraphics[width=\columnwidth]{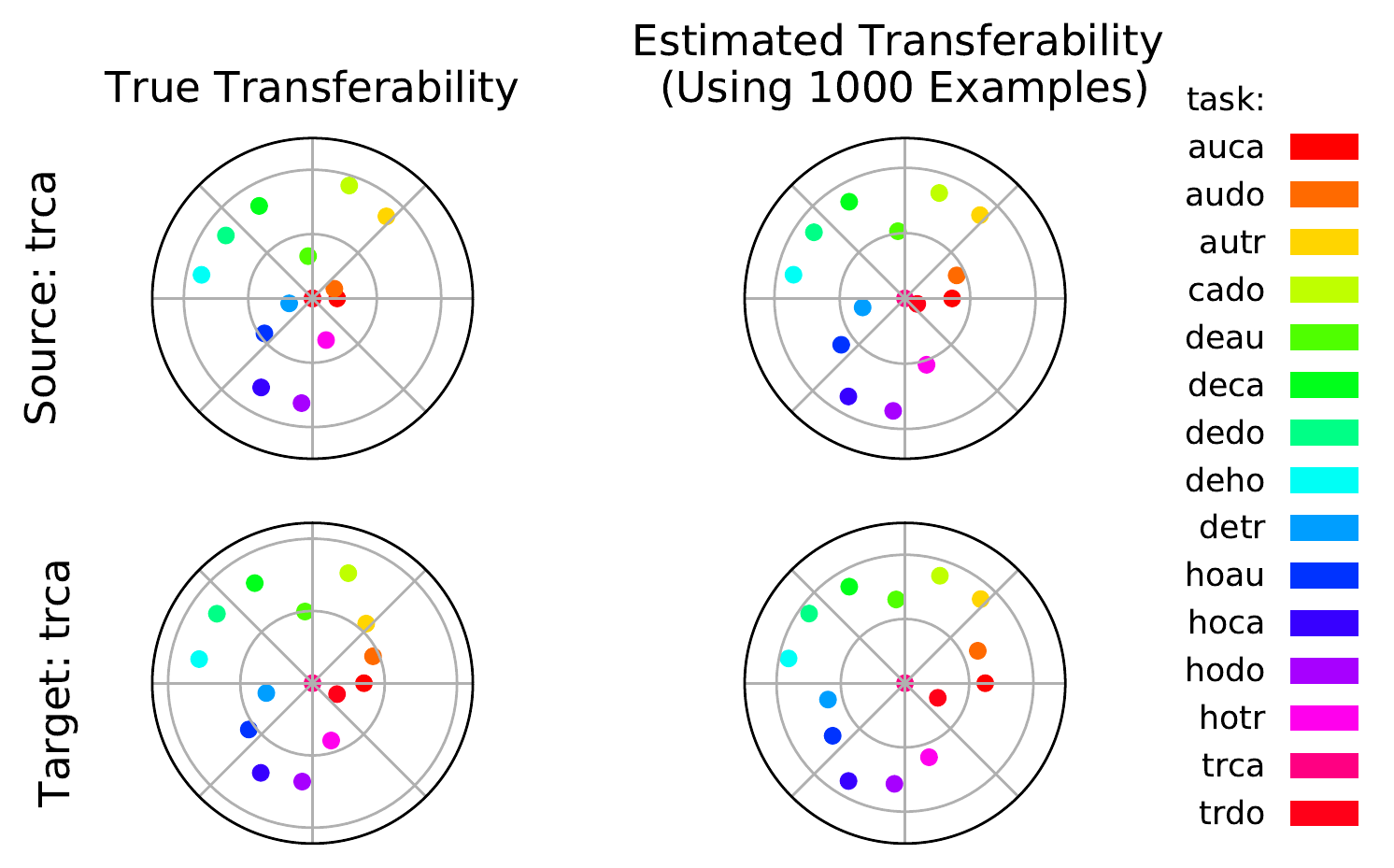}
    }
    \caption{
        Additional figures with other tasks as the source/target task, supplementing Fig.~\ref{fig6}.
    }
    \label{fig7}
\end{figure*}
\end{document}